# Coherent Integration of Databases by Abductive Logic Programming

**Ofer Arieli** OARIELI@MTA.AC.IL
*Department of Computer Science, The Academic College of Tel-Aviv,*
*4 Antokolski street, Tel-Aviv 61161, Israel.*

**Marc Denecker** MARC.DENECKER@CS.KULEUVEN.AC.BE
**Bert Van Nuffelen** BERT.VANNUFFELEN@CS.KULEUVEN.AC.BE
**Maurice Bruynooghe** MAURICE.BRUYNOOGHE@CS.KULEUVEN.AC.BE
*Department of Computer Science, Katholieke Universiteit Leuven,*
*Celestijnenlaan 200A, B-3001 Heverlee, Belgium.*

## Abstract

We introduce an abductive method for a coherent integration of independent data-sources. The idea is to compute a list of data-facts that should be inserted to the amalgamated database or retracted from it in order to restore its consistency. This method is implemented by an abductive solver, called $\mathcal{A}$system, that applies SLDNFA-resolution on a meta-theory that relates different, possibly contradicting, input databases. We also give a pure model-theoretic analysis of the possible ways to 'recover' consistent data from an inconsistent database in terms of those models of the database that exhibit as minimal inconsistent information as reasonably possible. This allows us to characterize the 'recovered databases' in terms of the 'preferred' (i.e., most consistent) models of the theory. The outcome is an abductive-based application that is sound and complete with respect to a corresponding model-based, preferential semantics, and – to the best of our knowledge – is more expressive (thus more general) than any other implementation of coherent integration of databases.

## 1. Introduction

Complex reasoning tasks often have to integrate information that is coming from different sources. One of the major challenges with this respect is to compose contradicting sources of information such that what is obtained would properly reflect the combination of the data-sources on one hand[1], and would still be *coherent* (in terms of consistency) on the other hand. There are a number of different issues involved in this process, the most important of which are the following:

1. Unification of the different ontologies and/or database schemas, in order to get a fixed (global) schema, and a translation of the integrity constraints[2] of each database to the new ontology.

2. Unification of translated integrity constraints in a single global set of integrity constraints. This means, in particular, elimination of contradictions among the translated

1. This property is sometimes called *compositionality* (Verbaeten, Denecker, & De Schreye, 1997, 2000).
2. I.e., the rules that represent intentional truths of a database domain.

integrity constraints, and inclusion of any global integrity constraint that is imposed on the integration process.

3. Integration of databases w.r.t. the unified set of integrity constraints, computed according to the previous item.

Each one of the issues mentioned above has its own difficulties and challenges. For instance, the first issue is considered, e.g., by Ullman (2000) and Lenzerini (2001, 2002), where questions such as how to express the relations between the 'global database schema' and the source (local) schemas, and how this influences query processing with respect to the global schema (Bertossi, Chomicki, Cortés, & Gutierrez, 2002), are dealt with[3].

The second issue above is concerned with the construction of a single, classically consistent, set of integrity constraints, applied on the integrated data. In database context, it is common to assume that such a set is pre-defined, and consists of global integrity constraints that are imposed on the integration process itself (Bertossi et al., 2002; Lenzerini, 2002). In such case there is no need to derive these constraints from the local databases. When different integrity constraints are specified in different local databases, it is required to integrate not only the database instances (as specified in issue 3 above), but also the integrity constraints themselves (issue 2). The reason for separating these two topics is that integrity constraints represent truths that should be valid in all situations, while a database instance exhibits an extensional truth, i.e., an actual situation. Consequently, the policy of resolving contradictions among integrity constraints is often different than the one that is applied on database facts, and often the former is applied before the latter.

Despite their different nature, both issues are based on some formalisms that maintain contradictions and allow to draw plausible conclusions from inconsistent situations. Roughly, there are two approaches to handle this problem:

- *Paraconsistent* formalisms, in which the amalgamated data may remain inconsistent, but the set of conclusions implied by it is not explosive, i.e.: not every fact follows from an inconsistent database, and so the inference process does not become trivial in the presence of contradictions. Paraconsistent procedures for integrating data, like those of Subrahmanian (1994) and de Amo, Carnielli, and Marcos (2002), are often based on a paraconsistent reasoning systems, such as LFI (Carnielli & Marcos, 2001), annotated logics (Subrahmanian, 1990; Kifer & Lozinskii, 1992; Arenas, Bertossi, & Kifer, 2000), or other non-classical proof procedures (Priest, 1991; Arieli & Avron, 1996; Avron, 2002; Carnielli & Marcos, 2002)[4].

- *Coherent* (consistency-based) methods, in which the amalgamated data is revised in order to restore consistency (see, e.g., Baral, Kraus, & Minker, 1991; Baral, Kraus, Minker, & Subrahmanain, 1992; Benferhat, Dubois, & Prade, 1995; Arenas, Bertossi,

3. For surveys on schema matching and related aspects, see also (Batini, Lenzerini, & Navathe, 1986) and (Rahm & Bernstein, 2001).
4. See also (Decker, 2003) for a historical perspective and some computational remarks on this kind of formalisms.

& Chomicki, 1999; Arieli & Avron, 1999; Greco & Zumpano, 2000; Liberatore & Schaerf, 2000; Bertossi & Schwind, 2002; Arieli, Denecker, Van Nuffelen, & Bruynooghe, 2004). In many cases the underlying formalisms of these approaches are closely related to the theory of belief revision (Alchourrón, Gärdenfors, & Makinson, 1995; Gärdenfors & Rott, 1995). In the context of database systems the idea is to consider consistent databases that are 'as close as possible' to the original database. These 'repaired' instances of the 'spoiled' database correspond to plausible and compact ways of restoring consistency.

In this paper we follow the latter approach, and consider abductive approaches that handle the third issue above, namely: coherent methods for integrating different data-sources (with the same ontology) w.r.t. a consistent set of integrity constraints[5]. The main difficulty in this process stems from the fact that even when each local database is consistent, the collective information of all the data-sources may not remain consistent anymore. In particular, facts that are specified in a particular database may violate some integrity constraints defined elsewhere, and so this data might contradict some elements in the unified set of integrity constraints. Moreover, as noted e.g. in (Lenzerini, 2001; Cali, Calvanese, De Giacomo, & Lenzerini, 2002), the ability to handle, in a plausible way, incomplete and inconsistent data, is an inherent property of any system for data integration with integrity constrains, no matter which integration phase is considered. Providing proper ways of gaining this property is a major concern here as well.

Our goal is therefore to find ways to properly 'repair' a combined (unified) database, and restore its consistency. For this, we consider a pure declarative representation of the composition of distributed data by a meta-theory, relating a number of different input databases (that may contradict each other) with a consistent output database. The underlying language of the theory is that of abductive logic programming (Kakas, Kowalski, & Toni, 1992; Denecker & Kakas, 2000). For reasoning with such theories we use an abductive system, called $\mathcal{A}$system (Kakas, Van Nuffelen, & Denecker, 2001; Van Nuffelen & Kakas, 2001), which is an abductive solver implementing SLDNFA-resolution (Denecker & De Schreye, 1992, 1998). The composing system is implemented by abductive reasoning on the meta-theory. In the context of this work, we have extended this system with an optimizing component that allows us to compute preferred coherent ways to restore the consistency of a given database. The system that is obtained induces an *operational semantics* for database integration. In the sequel we also consider some model-theoretic aspects of the problem, and define a *preferential semantics* (Shoham, 1988) for it. According to this semantics, the repaired databases are characterized in terms of the preferred models (i.e., the most-consistent valuations) of the underlying theory. We relate these approaches by showing that the $\mathcal{A}$system is sound and complete w.r.t. the model-based semantics. It is also noted that our framework supports reasoning with various types of special information, such as timestamps and source identification. Some implementation issues and experimen-

5. In this sense, one may view this work as a method for restoring the consistency of a single inconsistent database. We prefer, however, to treat it as an integration process of multiple sources, since it also has some mediating capabilities, such as source identification, making priorities among different data-sources, etc. (see, e.g., Section 4.6).

tal results are discussed as well.

The rest of this paper is organized as follows: in the next section we formally define our goal, namely: a coherent way to integrate different data-sources. In Section 3 we set up a semantics for this goal in terms of a corresponding model theory. Then, in Section 4 we introduce our abductive-based application for database integration. This is the main section of this paper, in which we also describe how a given integration problem can be represented in terms of meta logic programs, show how to reason with these programs by abductive computational models, present some experimental results, consider proper ways of reasoning with several types of special data, and show that our application is sound and complete with respect to the model-based semantics, considered in Section 3. Section 5 contains an overview of some related works, and in Section 6 we conclude with some further remarks, open issues, and future work[6].

## 2. Coherent Integration of Databases

We begin with a formal definition of our goal. In this paper we assume that we have a first-order language $L$, based on a fixed database schema $S$, and a fixed domain $D$. Every element of $D$ has a unique name. A *database instance* $\mathcal{D}$ consists of atoms in the language $L$ that are instances of the schema $S$. As such, every instance $\mathcal{D}$ has a finite active domain, which is a subset of $D$.

**Definition 1** A *database* is a pair $(\mathcal{D}, \mathcal{IC})$, where $\mathcal{D}$ is a database instance, and $\mathcal{IC}$, the set of *integrity constraints*, is a finite and classically consistent set of formulae in $L$.

Given a database $\mathcal{DB} = (\mathcal{D}, \mathcal{IC})$, we apply to it the closed word assumption, so only the facts that are explicitly mentioned in $\mathcal{D}$ are considered true. The underlying semantics corresponds, therefore, to minimal Herbrand interpretations.

**Definition 2** The *minimal Herbrand model* $\mathcal{H}^{\mathcal{D}}$ of a database instance $\mathcal{D}$ is the model of $\mathcal{D}$ that assigns true to all the ground instances of atomic formulae in $\mathcal{D}$, and false to all the other atoms.

There are different views on a database. One view is that it is a logic theory consisting of atoms and, implicitly, the closed world assumption (CWA) that indicates that all atoms not in the database are false. Another common view of a database is that it is a structure that consists of a certain domain and corresponding relations, representing the state of the world. Whenever there is a complete knowledge and all true atoms are represented in the database, both views coincide: the unique Herbrand model of the theory is the intended structure. However, in the context of independent data-sources, the assumption that each local database represents the state of the world is obviously false. However, we can still view a local database as an incomplete theory, and so treating a database as a theory rather than as a structure is more appropriate in our case.

6. This is a combined and extended version of (Arieli, Van Nuffelen, Denecker, & Bruynooghe, 2001) and (Arieli, Denecker, Van Nuffelen, & Bruynooghe, 2002).

**Definition 3** A formula $\psi$ *follows* from a database instance $\mathcal{D}$ (alternatively, $\mathcal{D}$ *entails* $\psi$; notation: $\mathcal{D} \models \psi$) if the minimal Herbrand model of $\mathcal{D}$ is also a model of $\psi$.

**Definition 4** A database $\mathcal{DB}=(\mathcal{D}, \mathcal{IC})$ is *consistent* if every formula in $\mathcal{IC}$ follows from $\mathcal{D}$ (notation: $\mathcal{D} \models \mathcal{IC}$).

Our goal is to integrate $n$ consistent local databases, $\mathcal{DB}_i=(\mathcal{D}_i, \mathcal{IC}_i)$ $(i=1,\ldots n)$ to one *consistent* database that contains as much information as possible from the local databases. The idea, therefore, is to consider the union of the distributed data, and then to restore its consistency in such a way that as much information as possible will be preserved.

**Notation 1** Let $\mathcal{DB}_i = (\mathcal{D}_i, \mathcal{IC}_i)$, $i = 1,\ldots n$, and let $\mathcal{I}(\mathcal{IC}_1,\ldots,\mathcal{IC}_n)$ be a classically consistent set of integrity constraints. We denote:

$$\mathcal{UDB} = (\bigcup_{i=1}^{n} \mathcal{D}_i \,,\, \mathcal{I}(\mathcal{IC}_1,\ldots,\mathcal{IC}_n)).$$

In the notation above, $\mathcal{I}$ is an operator that combines the integrity constraints and eliminates contradictions (see, e.g., Alferes, Leite, Pereira, & Quaresma, 2000; Alferes, Pereira, Przymusinska, & Przymusinski, 2002). As we have already noted, how to choose this operator and how to apply it on a specific database is beyond the scope of this paper. In cases that the union of all the integrity constraints is classically consistent, it makes sense to take $\mathcal{I}$ as the union operator. Global consistency of the integrity constraints is indeed a common assumption in the database literature (Arenas et al., 1999; Greco & Zumpano, 2000; Greco, Greco, & Zumpano, 2001; Bertossi et al., 2002; Konieczny & Pino Pérez, 2002; Lenzerini, 2002), but for the discussion here it is possible to take, instead of the union, any operator $\mathcal{I}$ for consistency restoration.

A key notion in database integration is the following:

**Definition 5** A *repair* of a database $\mathcal{DB}=(\mathcal{D}, \mathcal{IC})$ is a pair $(\mathsf{Insert}, \mathsf{Retract})$, such that:

1. $\mathsf{Insert} \cap \mathcal{D}=\emptyset$,
2. $\mathsf{Retract} \subseteq \mathcal{D}$[7],
3. $(\mathcal{D} \cup \mathsf{Insert} \setminus \mathsf{Retract}, \mathcal{IC})$ is a consistent database.

Intuitively, $\mathsf{Insert}$ is a set of elements that should be inserted into $\mathcal{D}$ and $\mathsf{Retract}$ is a set of elements that should be removed from $\mathcal{D}$ in order to have a consistent database.

As noted above, repair of a given database is a key notion in many formalisms for data integration. In the context of database systems, this notion was first introduced by Arenas, Bertossi, and Chomicki (1999), and later considered by many others (e.g., Greco & Zumpano, 2000; Liberatore & Schaerf, 2000; Franconi, Palma, Leone, Perri, & Scarcello, 2001; Bertossi et al., 2002; Bertossi & Schwind, 2002; de Amo et al., 2002; Arenas, Bertossi, & Chomicki, 2003; Arieli et al., 2004). Earlier versions of repairs and inclusion-based consistency restoration may be traced back to Dalal (1988) and Winslett (1988).

7. Note that by conditions (1) and (2) it follows that $\mathsf{Insert} \cap \mathsf{Retract}=\emptyset$.

**Definition 6** A *repaired database* of $\mathcal{DB} = (\mathcal{D}, \mathcal{IC})$ is a consistent database of the form $(\mathcal{D} \cup \mathsf{Insert} \setminus \mathsf{Retract}\,, \mathcal{IC})$, where $(\mathsf{Insert}, \mathsf{Retract})$ is a repair of $\mathcal{DB}$.

As there may be many ways to repair an inconsistent database[8], it is often convenient to make preferences among the possible repairs, and consider only the most preferred ones. Below are two common preference criteria for preferring a repair $(\mathsf{Insert}, \mathsf{Retract})$ over a repair $(\mathsf{Insert}', \mathsf{Retract}')$:

**Definition 7** Let $(\mathsf{Insert}, \mathsf{Retract})$ and $(\mathsf{Insert}', \mathsf{Retract}')$ be two repairs of a given database.

- *set inclusion preference criterion* : $(\mathsf{Insert}', \mathsf{Retract}') \leq_i (\mathsf{Insert}, \mathsf{Retract})$, if $\mathsf{Insert} \subseteq \mathsf{Insert}'$ and $\mathsf{Retract} \subseteq \mathsf{Retract}'$.
- *minimal cardinality preference criterion*: $(\mathsf{Insert}', \mathsf{Retract}') \leq_c (\mathsf{Insert}, \mathsf{Retract})$, if $|\mathsf{Insert}| + |\mathsf{Retract}| \leq |\mathsf{Insert}'| + |\mathsf{Retract}'|$.

Set inclusion is also considered in (Arenas et al., 1999; Greco & Zumpano, 2000; Bertossi et al., 2002; Bertossi & Schwind, 2002; de Amo et al., 2002; Arenas et al., 2003; Arieli et al., 2004, and others), minimal cardinality is considered, e.g., in (Dalal, 1988; Liberatore & Schaerf, 2000; Arenas et al., 2003; Arieli et al., 2004).

In what follows we assume that the preference relation $\leq$ is a fixed pre-order that represents some preference criterion on the set of repairs (and we shall omit subscript notations in it whenever possible). We shall also assume that if $(\emptyset, \emptyset)$ is a valid repair, it is the $\leq$-least (i.e., the 'best') one. This corresponds to the intuition that a database should not be repaired unless it is inconsistent.

**Definition 8** A $\leq$*-preferred repair* of $\mathcal{DB}$ is a repair $(\mathsf{Insert}, \mathsf{Retract})$ of $\mathcal{DB}$, s.t. for every repair $(\mathsf{Insert}', \mathsf{Retract}')$ of $\mathcal{DB}$, if $(\mathsf{Insert}, \mathsf{Retract}) \leq (\mathsf{Insert}', \mathsf{Retract}')$ then $(\mathsf{Insert}', \mathsf{Retract}') \leq (\mathsf{Insert}, \mathsf{Retract})$. The set of all the $\leq$-preferred repairs of $\mathcal{DB}$ is denoted by $!(\mathcal{DB}, \leq)$.

**Definition 9** A $\leq$*-repaired database* of $\mathcal{DB}$ is a repaired database of $\mathcal{DB}$, constructed from a $\leq$-preferred repair of $\mathcal{DB}$. The set of all the $\leq$-repaired databases is denoted by:

$$\mathcal{R}(\mathcal{DB}, \leq) = \{\, (\mathcal{D} \cup \mathsf{Insert} \setminus \mathsf{Retract}\,, \mathcal{IC}) \mid (\mathsf{Insert}, \mathsf{Retract}) \in \,!(\mathcal{DB}, \leq) \,\}.$$

Note that if $\mathcal{DB}$ is consistent and $\leq$ is a preference relation, then $\mathcal{DB}$ is the only $\leq$-repaired database of itself (thus, there is nothing to repair in this case, as expected).

**Note 1** It is usual to refer to the $\leq$-preferred databases of $\mathcal{DB}$ as the consistent databases that are 'as close as possible' to $\mathcal{DB}$ itself (see, e.g., Arenas, Bertossi, & Chomicki, 1999; Liberatore & Schaerf, 2000; de Amo, Carnielli, & Marcos, 2002; Konieczny & Pino Pérez, 2002; Arenas, Bertossi, & Chomicki, 2003; Arieli, Denecker, Van Nuffelen, & Bruynooghe, 2004). Indeed, let

$$\mathtt{dist}(\mathcal{D}_1, \mathcal{D}_2) = (\mathcal{D}_1 \setminus \mathcal{D}_2) \cup (\mathcal{D}_2 \setminus \mathcal{D}_1).$$

8. Some repairs may be trivial and/or useless, though. For instance, one way to eliminate the inconsistency in $(\mathcal{D}, \mathcal{IC}) = (\{p, q, r\}, \{\neg p\})$ is by deleting every element in $\mathcal{D}$, but this is certainly not the optimal way of restoring consistency in this case.

It is easy to see that $\mathcal{DB}' = (\mathcal{D}', \mathcal{IC})$ is a $\leq_i$-repaired database of $\mathcal{DB} = (\mathcal{D}, \mathcal{IC})$, if the set $\mathtt{dist}(\mathcal{D}', \mathcal{D})$ is minimal (w.r.t. set inclusion) among all the sets of the form $\mathtt{dist}(\mathcal{D}'', \mathcal{D})$, where $\mathcal{D}'' \models \mathcal{IC}$. Similarly, if $|S|$ denotes the size of $S$, then $\mathcal{DB}' = (\mathcal{D}', \mathcal{IC})$ is a $\leq_c$-repaired database of $\mathcal{DB} = (\mathcal{D}, \mathcal{IC})$, if $|\mathtt{dist}(\mathcal{D}', \mathcal{D})| = \min\{|\mathtt{dist}(\mathcal{D}'', \mathcal{D})| \mid \mathcal{D}'' \models \mathcal{IC}\}$.

Given $n$ databases and a preference criterion $\leq$, our goal is therefore to compute the set $\mathcal{R}(\mathcal{UDB}, \leq)$ of the $\leq$-repaired databases of the unified database, $\mathcal{UDB}$ (Notation 1). The reasoner may use different strategies to determine the consequences of this set. Among the common approaches are the skeptical (conservative) one, that it is based on a 'consensus' among all the elements of $\mathcal{R}(\mathcal{UDB}, \leq)$ (see Arenas et al., 1999; Greco & Zumpano, 2000), a 'credulous' approach, in which entailments are determined by any element in $\mathcal{R}(\mathcal{UDB}, \leq)$, an approach that is based on a 'majority vote' (Lin & Mendelzon, 1998; Konieczny & Pino Pérez, 2002), etc. In cases where processing time is a major consideration, one may want to speed-up the computations by considering *any* repaired database. In such cases it is sufficient to find an arbitrary element in the set $\mathcal{R}(\mathcal{UDB}, \leq)$.

Below are some examples[9] of the integration process[10].

**Example 1** Consider a relation *teaches* of the schema (`course_name`, `teacher_name`), and an integrity constraint, stating that the same course cannot be taught by two different teachers:

$$\mathcal{IC} = \{\, \forall X \forall Y \forall Z\, (teaches(X,Y) \wedge teaches(X,Z) \;\rightarrow\; Y = Z) \,\}.$$

Consider now the following two databases:

$$\mathcal{DB}_1 = (\, \{teaches(c_1, n_1),\; teaches(c_2, n_2)\},\, \mathcal{IC}\,),$$
$$\mathcal{DB}_2 = (\, \{teaches(c_2, n_3)\},\, \mathcal{IC}).$$

Clearly, the unified database $\mathcal{DB}_1 \cup \mathcal{DB}_2$ is inconsistent. It has two preferred repairs, which are $(\emptyset,\, \{teaches(c_2, n_3)\})$ and $(\emptyset,\, \{teaches(c_2, n_2)\})$. The corresponding repaired databases are the following:

$$\mathcal{R}_1 = (\, \{teaches(c_1, n_1),\; teaches(c_2, n_2)\},\, \mathcal{IC}\,),$$
$$\mathcal{R}_2 = (\, \{teaches(c_1, n_1),\; teaches(c_2, n_3)\},\, \mathcal{IC}\,).$$

Thus, e.g., $teaches(c_1, n_1)$ is true both in the conservative approach and the credulous approach to database integration, while the conclusion $teaches(c_2, n_2)$ is supported only by credulous reasoning.

**Example 2** Consider databases with relations *class* and *supply*, of schemas (`item`, `type`) and (`supplier`, `department`, `item`), respectively. Let

$$\mathcal{DB}_1 = (\{supply(c_1, d_1, i_1),\; class(i_1, t_1)\},\, \mathcal{IC}),$$
$$\mathcal{DB}_2 = (\{supply(c_2, d_2, i_2),\; class(i_2, t_1)\},\, \emptyset),$$

where $\mathcal{IC} = \{\, \forall X \forall Y \forall Z\, (supply(X,Y,Z) \wedge class(Z, t_1) \rightarrow X = c_1) \,\}$ states that only supplier

9. See, e.g., (Arenas et al., 1999; Greco & Zumpano, 2000; Bertossi & Schwind, 2002) for further discussions on these examples.
10. In all the following examples we use set inclusion as the preference criterion, and take the operator $\mathcal{I}$ that combines integrity constraints (see Notation 1) to be the union operator.

$c_1$ can supply items of type $t_1$. Again, $\mathcal{DB}_1 \cup \mathcal{DB}_2$ is inconsistent, and has two preferred repairs: $(\emptyset,\ \{supply(c_2, d_2, i_2)\})$ and $(\emptyset,\ \{class(i_2, t_1)\})$. It follows that there are two repairs of this database:

$$\mathcal{R}_1 = (\,\{supply(c_1, d_1, i_1),\ class(i_1, t_1),\ class(i_2, t_1)\},\ \mathcal{IC}\,),$$
$$\mathcal{R}_2 = (\,\{supply(c_1, d_1, i_1),\ supply(c_2, d_2, i_2),\ class(i_1, t_1)\},\ \mathcal{IC}\,).$$

**Example 3** Let $\mathcal{D}_1 = \{p(a),\ p(b)\}$, $\mathcal{D}_2 = \{q(a),\ q(c)\}$, and $\mathcal{IC} = \{\forall X(p(X) \rightarrow q(X))\}$. Again, $(\mathcal{D}_1, \emptyset) \cup (\mathcal{D}_2, \mathcal{IC})$ is inconsistent. The corresponding preferred repairs are $(\{q(b)\},\ \emptyset)$ and $(\emptyset,\ \{p(b)\})$. Thus, the repaired databases are the following:

$$\mathcal{R}_1 = (\,\{p(a),\ p(b),\ q(a),\ q(b),\ q(c)\},\ \mathcal{IC}\,),$$
$$\mathcal{R}_2 = (\,\{p(a),\ q(a),\ q(c)\},\ \mathcal{IC}\,).$$

In this case, then, both the 'consensus approach' and the 'credulous approach' allow to infer, e.g., that $p(a)$ holds, while $p(b)$ is supported only by credulous reasoning, and $p(c)$ is not supported by either of these approaches.

## 3. Model-based Characterization of Repairs

In this section we set up a semantics for describing repairs and preferred repairs in terms of a corresponding model theory. This will allow us, in particular, to give an alternative description of preferred repairs, this time in terms of a preferential semantics for database theory.

As database semantics is usually defined in terms of two-valued (Herbrand) models (cf. Definition 2 and the discussion that proceeds it), it is natural to consider two-valued semantics first. We show that arbitrary repairs can be represented by *two-valued* models of the integrity constraints. When a database is inconsistent, then by definition, there is no two-valued interpretation which satisfies both its database instance and its integrity constraints. A standard way to cope with this type of inconsistencies is to move to *multiple-valued semantics* for reasoning with inconsistent and incomplete information (see, e.g., Subrahmanian, 1990, 1994; Messing, 1997; Arieli & Avron, 1999; Arenas, Bertossi, & Kifer, 2000; de Amo, Carnielli, & Marcos, 2002). What we will show below, is that repairs can be characterized by three-valued models of the whole database, that is, of the database instance and the integrity constraints. Finally, we concentrate on the most preferred repairs, and show that a certain subset of the three-valued models can be used for characterizing $\leq$-preferred repairs.

**Definition 10** Given a valuation $\nu$ and a truth value $x$. Denote:

$$\nu^x = \{p \mid p \text{ is an atomic formula, and } \nu(p) = x\}^{11}.$$

The following two propositions characterize repairs in terms of two-valued structures.

**Proposition 1** *Let $(\mathcal{D}, \mathcal{IC})$ be a database and let $M$ be a two-valued model of $\mathcal{IC}$. Let $\mathsf{Insert} = M^t \setminus \mathcal{D}$ and $\mathsf{Retract} = \mathcal{D} \setminus M^t$. Then $(\mathsf{Insert}, \mathsf{Retract})$ is a repair of $(\mathcal{D}, \mathcal{IC})$.*

11. Note, in particular, that in terms of Definition 2, if $\nu = \mathcal{H}^{\mathcal{D}}$ and $x = t$, we have that $\nu^x = \mathcal{D}$.

*Proof:* The definitions of $\mathsf{Insert}$ and $\mathsf{Retract}$ immediately imply that $\mathsf{Insert} \cap \mathcal{D} = \emptyset$ and $\mathsf{Retract} \subseteq \mathcal{D}$. For the last condition in Definition 5, note that $\mathcal{D} \cup \mathsf{Insert} \setminus \mathsf{Retract} = \mathcal{D} \cup (M^t \setminus \mathcal{D}) \setminus (\mathcal{D} \setminus M^t) = M^t$. It follows that $M$ is the least Herbrand model of $\mathcal{D} \cup \mathsf{Insert} \setminus \mathsf{Retract}$ and it is also a model of $\mathcal{IC}$, therefore $\mathcal{D} \cup \mathsf{Insert} \setminus \mathsf{Retract} \models \mathcal{IC}$. □

**Proposition 2** *Let* $(\mathsf{Insert}, \mathsf{Retract})$ *be a repair of a database* $(\mathcal{D}, \mathcal{IC})$. *Then there is a two-valued model* $M$ *of* $\mathcal{IC}$ *such that* $\mathsf{Insert} = M^t \setminus \mathcal{D}$ *and* $\mathsf{Retract} = \mathcal{D} \setminus M^t$.

*Proof:* Consider a valuation $M$, defined for every atom $p$ as follows:

$$M(p) = \begin{cases} t & \text{if } p \in \mathcal{D} \cup \mathsf{Insert} \setminus \mathsf{Retract}, \\ f & \text{otherwise.} \end{cases}$$

By its definition, $M$ is a minimal Herbrand model of $\mathcal{D} \cup \mathsf{Insert} \setminus \mathsf{Retract}$. Now, since $(\mathsf{Insert}, \mathsf{Retract})$ is a repair of $(\mathcal{D}, \mathcal{IC})$, we have that $\mathcal{D} \cup \mathsf{Insert} \setminus \mathsf{Retract} \models \mathcal{IC}$, thus $M$ is a (two-valued) model of $\mathcal{IC}$. Moreover, since $(\mathsf{Insert}, \mathsf{Retract})$ is a repair, necessarily $\mathsf{Insert} \cap \mathcal{D} = \emptyset$ and $\mathsf{Retract} \subseteq \mathcal{D}$, hence we have the following:

- $M^t \setminus \mathcal{D} = (\mathcal{D} \cup \mathsf{Insert} \setminus \mathsf{Retract}) \setminus \mathcal{D} = \mathsf{Insert}$,
- $\mathcal{D} \setminus M^t = \mathcal{D} \setminus (\mathcal{D} \cup \mathsf{Insert} \setminus \mathsf{Retract}) = \mathsf{Retract}$. □

When a database is inconsistent, it has no models that satisfy both its integrity constraints and its database instance. One common method to overcome such an inconsistency is to introduce additional truth-values that intuitively represent partial knowledge, different amounts of beliefs, etc. (see, e.g., Priest, 1989, 1991; Subrahmanian, 1990; Fitting, 1991; Arieli, 1999; Arenas et al., 2000; Avron, 2002). Here we follow this guideline, and consider database integration in the context of a three-valued semantics. The benefit of this is that, as we show below, *any* database has some three-valued models, from which it is possible to pinpoint the inconsistent information, and accordingly construct repairs.

The underlying three-valued semantics considered here is induced by the algebraic structure $\mathcal{THREE}$, shown in the double-Hasse diagram of Figure 1[12].

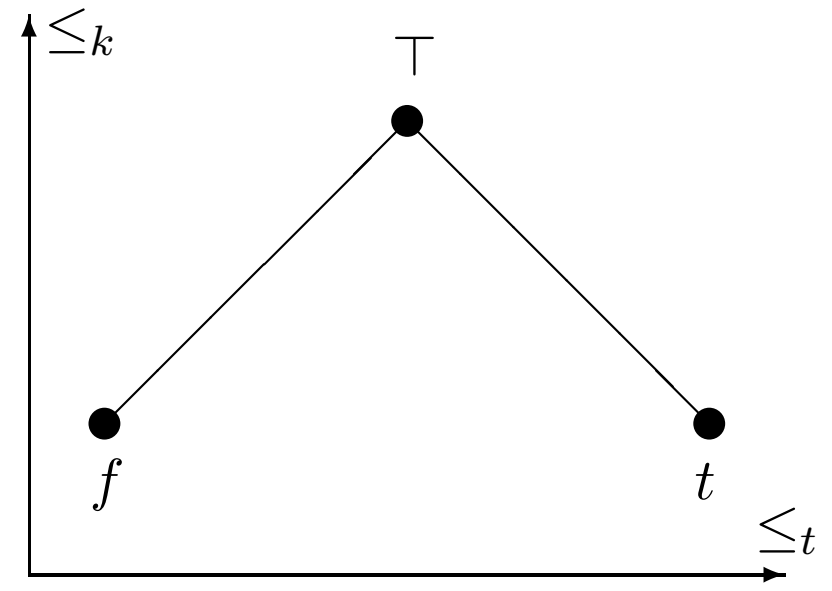

Figure 1: The structure $\mathcal{THREE}$

Intuitively, the elements $t$ and $f$ in $\mathcal{THREE}$ correspond to the usual classical values `true` and `false`, while the third element, $\top$, represents inconsistent information (or belief).

12. This structure is used for reasoning with inconsistency by several other three-valued formalisms, such as LFI (Carnielli & Marcos, 2001, 2002) and LP (Priest, 1989, 1991).

Viewed horizontally, $\mathcal{THREE}$ is a complete lattice. According to this view, $f$ is the minimal element, $t$ is the maximal one, and $\top$ is an intermediate element. The corresponding order relation, $\leq_t$, intuitively represents differences in the amount of *truth* that each element exhibits. We denote the meet, join, and the order reversing operation on $\leq_t$ by $\wedge$, $\vee$, and $\neg$ (respectively). Viewed vertically, $\mathcal{THREE}$ is a semi-upper lattice. In this view, $\top$ is the maximal element and the two 'classical values' are incomparable. This partial order, denoted by $\leq_k$, may be intuitively understood as representing differences in the amount of *knowledge* (or information) that each element represents[13]. We denote by $\oplus$ the join operation on $\leq_k$[14].

Various semantic notions can be defined on $\mathcal{THREE}$ as natural generalizations of similar classical ones: a *valuation* $\nu$ is a function that assigns a truth value in $\mathcal{THREE}$ to each atomic formula. Given a valuation $\nu$, truth values $x_i \in \{t, f, \top\}$, and atomic formulae $p_i$, we shall sometimes write $\nu = \{p_i : x_i\}$ instead of $\nu(p_i) = x_i \quad (i = 1, 2 \ldots)$. Any valuation is extended to complex formulae in the obvious way. For instance, $\nu(\neg\psi) = \neg\nu(\psi)$, $\nu(\psi \wedge \phi) = \nu(\psi) \wedge \nu(\psi)$, and so forth[15]. The set of the *designated* truth values in $\mathcal{THREE}$ (i.e., those elements in $\mathcal{THREE}$ that represent true assertions) consists of $t$ and $\top$. A valuation $\nu$ *satisfies* a formula $\psi$ iff $\nu(\psi)$ is designated. A valuation that assigns a designated value to every formula in a theory $\mathcal{T}$ is a (three-valued) *model* of $\mathcal{T}$.

**Lemma 1** *Let $\nu$ and $\mu$ be three-valued valuations s.t. for every atom $p$, $\nu(p) \geq_k \mu(p)$. Then for every formula $\psi$, $\nu(\psi) \geq_k \mu(\psi)$.*

*Proof:* By induction on the structure of $\psi$. □

We shall write $\nu \geq_k \mu$, if $\nu$ and $\mu$ are three-valued valuations, for which the condition of Lemma 1 holds.

**Lemma 2** *If $\nu \geq_k \mu$ and $\mu$ is a model of some theory $\mathcal{T}$, then $\nu$ is also a model of $\mathcal{T}$.*

*Proof:* For every formula $\psi \in \mathcal{T}$, $\mu(\psi)$ is designated. Hence, by Lemma 1, for every formula $\psi \in \mathcal{T}$ $\nu(\psi)$ is also designated, and so $\nu$ is a model of $\mathcal{T}$. □

Next we characterize the repairs of a database $\mathcal{DB}$ by its three-valued models:

**Proposition 3** *Let $(\mathcal{D}, \mathcal{IC})$ be a database and let $M$ be a two-valued model of $\mathcal{IC}$. Consider the three-valued valuation $N$, defined for every atom $p$ by $N(p) = \mathcal{H}^{\mathcal{D}}(p) \oplus M(p)$, and let* $\mathsf{Insert} = N^{\top} \setminus \mathcal{D}$, $\mathsf{Retract} = N^{\top} \cap \mathcal{D}$. *Then:*

1. *$N$ is a three-valued model of $\mathcal{D} \cup \mathcal{IC}$, and*

13. See (Belnap, 1977; Ginsberg, 1988; Fitting, 1991) for a more detailed discussion on these orders and their intuitive meaning.
14. We follow here the notations of Fitting (1990, 1991).
15. As usual, we use here the same logical symbol to denote the connective that appear on the left-hand side of an equation, and the corresponding operator on $\mathcal{THREE}$ that appear on the right-hand side of the same equation.

*2.* (Insert, Retract) *is a repair of* ($\mathcal{D}$, $\mathcal{IC}$).

Proposition 3 shows that repairs of a database ($\mathcal{D}$, $\mathcal{IC}$) may be constructed in a standard (uniform) way by considering three-valued models that are the $\leq_k$-least upper bounds of two two-valued valuations: the minimal Herbrand model of the database instance, and a two-valued model of the integrity constraints. Proposition 4 below shows that any repair of ($\mathcal{D}$, $\mathcal{IC}$) is of this form.

Before we give a proof for Proposition 3, let's demonstrate it by a simple example.

**Example 4** Let $\mathcal{DB} = (\{p, r\}\,,\, \{p \rightarrow q\})$. Then $\mathcal{H}^{\mathcal{D}} = \{p\!:\!t,\ q\!:\!f,\ r\!:\!t\}$, and the two-valued models of $\mathcal{IC} = \{p \rightarrow q\}$ are $\{p\!:\!t, q\!:\!t, r\!:\!t\}$, $\{p\!:\!t, q\!:\!t, r\!:\!f\}$, $\{p\!:\!f, q\!:\!t, r\!:\!t\}$, $\{p\!:\!f, q\!:\!t, r\!:\!f\}$, $\{p\!:\!f, q\!:\!f, r\!:\!t\}$, and $\{p\!:\!f, q\!:\!f, r\!:\!f\}$. Thus, the (three-valued) models of the form $\mathcal{H}^{\mathcal{D}} \oplus M$, where $M$ is a two-valued model of $\mathcal{IC}$, are $\{p\!:\!t, q\!:\!\top, r\!:\!t\}$, $\{p\!:\!t, q\!:\!\top, r\!:\!\top\}$, $\{p\!:\!\top, q\!:\!\top, r\!:\!t\}$, $\{p:\top,\ q:\top,\ r:\top\}$, $\{p:\top,\ q:f,\ r:t\}$ and $\{p:\top,\ q:f,\ r:\top\}$. By Proposition 3, the pairs $(\{q\}, \{\})$, $(\{q\}, \{r\})$, $(\{q\}, \{p\})$, $(\{q\}, \{p, r\})$, $(\{\}, \{p\})$, and $(\{\}, \{p, r\})$, are repairs of $\mathcal{DB}$.

*Proof of Proposition 3:* Since by the definition of $N$, $N \geq_k \mathcal{H}^{\mathcal{D}}$, and since $\mathcal{H}^{\mathcal{D}}$ is a model of $\mathcal{D}$, Lemma 2 implies that $N$ is a model $\mathcal{D}$. Similarly, $N \geq_k M$, and $M$ is a model of $\mathcal{IC}$, thus by the same lemma $N$ is also a model of $\mathcal{IC}$.

For the second part, we observe that $\mathsf{Insert} = N^{\top} \setminus \mathcal{D} = M^t \setminus \mathcal{D}$ and $\mathsf{Retract} = N^{\top} \cap \mathcal{D} = M^f \cap \mathcal{D} = \mathcal{D} \setminus M^t$. Now, $M$ is a two-valued model of $\mathcal{IC}$ and hence, by Proposition 1, (Insert, Retract) is a repair of ($\mathcal{D}$, $\mathcal{IC}$). □

Note that the specific form of the three-valued valuations considered in Proposition 3 is essential here, as the proposition does not hold for *every* three-valued model of $\mathcal{D} \cup \mathcal{IC}$. To see this consider, e.g., $\mathcal{D} = \{\}$, $\mathcal{IC} = \{p\ ,\ p \rightarrow q\}$, and a three valued valuation $N$ that assigns $\top$ to $p$ and $t$ to $q$. Clearly, $N$ is a model of $\mathcal{D} \cup \mathcal{IC}$, but the corresponding update, $(N^{\top} \setminus \mathcal{D}\ ,\ N^{\top} \cap \mathcal{D}) = (\{p\}, \{\})$ is *not* a repair of ($\mathcal{D}$, $\mathcal{IC}$), since $(\{p\}, \mathcal{IC})$ is not a consistent database.

Again, as we have noted above, it is possible to show that the converse of Proposition 3 is also true:

**Proposition 4** *Let* (Insert, Retract) *be a repair of a database* ($\mathcal{D}$, $\mathcal{IC}$). *Then there is a three-valued model $N$ of $\mathcal{D} \cup \mathcal{IC}$, such that* $\mathsf{Insert} = N^{\top} \setminus \mathcal{D}$ *and* $\mathsf{Retract} = N^{\top} \cap \mathcal{D}$.

*Proof:* Consider a valuation $N$, defined for every atom $p$ as follows:

$$N(p) = \begin{cases} \top & \text{if } p \in \mathsf{Insert} \cup \mathsf{Retract}, \\ t & \text{if } p \notin \mathsf{Insert} \cup \mathsf{Retract} \text{ but } p \in \mathcal{D}, \\ f & \text{otherwise.} \end{cases}$$

By the definition of $N$ and since (Insert, Retract) is a repair of ($\mathcal{D}$, $\mathcal{IC}$), we have that $N^{\top} \setminus \mathcal{D} = (\mathsf{Insert} \cup \mathsf{Retract}) \setminus \mathcal{D} = \mathsf{Insert}$ and $N^{\top} \cap \mathcal{D} = (\mathsf{Insert} \cup \mathsf{Retract}) \cap \mathcal{D} = \mathsf{Retract}$.

It remains to show that $N$ is a (three-valued) model of $\mathcal{D}$ and $\mathcal{IC}$. It is a three-valued model of $\mathcal{D}$ because for every $p \in \mathcal{D}$, $N(p) \in \{t, \top\}$. Regarding $\mathcal{IC}$, (Insert, Retract) is a repair of $(\mathcal{D}, \mathcal{IC})$, thus every formula in $\mathcal{IC}$ is true in the least Herbrand model $M$ of $\mathcal{D}' = \mathcal{D} \cup \mathsf{Insert} \setminus \mathsf{Retract}$. In particular, $M(q) = t$ for every $q \in \mathcal{D}'$. But since for every $p \in \mathcal{D} \cup \mathsf{Insert}$ we have that $N(p) \in \{t, \top\}$ and $\mathcal{D}' \subseteq \mathcal{D} \cup \mathsf{Insert}$, necessarily $\forall q \in \mathcal{D}'\ N(q) \in \{t, \top\}$. It follows that for every $q \in \mathcal{D}'$, $N(q) \geq_k M(q) = t$, thus by Lemma 1 and Lemma 2, $N$ must also be a (three-valued) model of $\mathcal{D}'$. Hence $N$ is a model of $\mathcal{IC}$. □

The last two propositions characterize the repairs of $\mathcal{UDB}$ in terms of pairs that are associated with certain three-valued models of $\mathcal{D} \cup \mathcal{IC}$. We shall denote the elements of these pairs as follows:

**Notation 2** Let $N$ be a three-valued model and let $\mathcal{DB} = (\mathcal{D}, \mathcal{IC})$ be a database. Denote: $\mathsf{Insert}^N = N^\top \setminus \mathcal{D}$ and $\mathsf{Retract}^N = N^\top \cap \mathcal{D}$.

We conclude this model-based analysis by characterizing the set of the $\leq$-preferred repairs, where $\leq$ is one of the preference criteria considered in Definition 7 (i.e., set inclusion or minimal cardinality). As the propositions below show, common considerations on how inconsistent databases can be 'properly' recovered (e.g., keeping the amount of changes as minimal as possible, being 'as close as possible' to the original instance, etc.) can be captured by preferential models in the context of preferential semantics (Shoham, 1988). The idea is to define some order relation on the set of the (three-valued) models of the database. This relation intuitively captures some criterion for making preferences among the relevant models. Then, only the 'most preferred' models (those that are minimal with respect to the underlying order relation) are considered in order to determine how the database should be repaired. Below we formalize this idea:

**Definition 11** Given a database $\mathcal{DB} = (\mathcal{D}, \mathcal{IC})$, denote:

$$\mathcal{M}^{\mathcal{DB}} = \{N \mid N \geq_k \mathcal{H}^{\mathcal{D}} \oplus M \text{ for some classical model } M \text{ of } \mathcal{IC}\}^{16}.$$

**Example 5** Consider again the database $\mathcal{DB} = (\{p, r\}\ ,\ \{p \to q\})$ of Example 4. As we have shown, there are six valuations of the form $\mathcal{H}^{\mathcal{D}} \oplus M$, for some two-valued model $M$ of $\mathcal{IC}$, namely:

$$\{p{:}t\ ,\ q{:}\top,\ r{:}t\}, \quad \{p{:}t\ ,\ q{:}\top,\ r{:}\top\}, \quad \{p{:}\top,\ q{:}\top,\ r{:}t\},$$
$$\{p{:}\top,\ q{:}\top,\ r{:}\top\}, \quad \{p{:}\top,\ q{:}f,\ r{:}t\}, \quad \{p{:}\top,\ q{:}f,\ r{:}\top\}.$$

The $k$-minimal models among these models are $\{p{:}t,\ q{:}\top,\ r{:}t\}$ and $\{p{:}\top,\ q{:}f,\ r{:}t\}$, thus $\mathcal{M}^{\mathcal{DB}} = \{N \mid N(p) \geq_k t,\ N(q) = \top,\ N(r) \geq_k t\} \cup \{N \mid N(p) = \top, N(q) \geq_k f,\ N(r) \geq_k t\}$.

Preference orders should reflect some normality considerations applied on the relevant set of valuations ($\mathcal{M}^{\mathcal{DB}}$, in our case); $\nu$ is preferable than $\mu$, if $\nu$ describes a situation that is more common (plausible) than the one described by $\mu$. Hence, a natural way to define preferences in our case is by minimizing inconsistencies. We thus get the following definition:

**Definition 12** Let $\mathcal{S}$ be a set of three-valued valuations, and $N_1, N_2 \in \mathcal{S}$.

16. Note that $N$ is a *three-valued* valuation and $M$ is a *two-valued* model of $\mathcal{IC}$.

- $N_1$ is $\leq_i$-*more consistent* than $N_2$, if $N_1^{\top} \subset N_2^{\top}$.
- $N_1$ is $\leq_c$-*more consistent* than $N_2$, if $|N_1^{\top}| < |N_2^{\top}|$.
- $N \in \mathcal{S}$ is $\leq_i$-*maximally consistent* in $\mathcal{S}$ (respectively, $N$ is $\leq_c$-*maximally consistent* in $\mathcal{S}$), if there is no $N' \in \mathcal{S}$ that is $\leq_i$-more consistent than $N$ (respectively, no $N' \in \mathcal{S}$ is $\leq_c$-more consistent than $N$).

The following propositions show that there is a close relationship between most consistent models of $\mathcal{M}^{\mathcal{DB}}$ and the preferred repairs of $\mathcal{DB}$.

**Proposition 5** *If $N$ is a $\leq_i$-maximally consistent element in $\mathcal{M}^{\mathcal{DB}}$, then* $(\mathsf{Insert}^N,\ \mathsf{Retract}^N)$ *is a $\leq_i$-preferred repair of $\mathcal{DB}$.*

*Proof:* By Proposition 3, $(\mathsf{Insert}^N,\ \mathsf{Retract}^N)$ is a repair of $\mathcal{DB}$. If it is not a $\leq_i$-preferred repair of $\mathcal{DB}$, then there is a repair $(\mathsf{Insert},\ \mathsf{Retract})$ s.t. $\mathsf{Insert} \subseteq \mathsf{Insert}^N$, $\mathsf{Retract} \subseteq \mathsf{Retract}^N$, and $\mathsf{Insert} \cup \mathsf{Retract} \subset \mathsf{Insert}^N \cup \mathsf{Retract}^N$. By Proposition 4 and its proof, there is an element $N' \in \mathcal{M}^{\mathcal{DB}}$ s.t. $\mathsf{Insert} = \mathsf{Insert}^{N'}$, $\mathsf{Retract} = \mathsf{Retract}^{N'}$, and $(N')^{\top} = \mathsf{Insert}^{N'} \cup \mathsf{Retract}^{N'}$. It follows, then, that $(N')^{\top} \subset N^{\top}$, and so $N$ is not a maximally consistent in $\mathcal{M}^{\mathcal{DB}}$, but this is a contradiction to the definition of $N$. □

**Proposition 6** *Suppose that* $(\mathsf{Insert},\ \mathsf{Retract})$ *is a $\leq_i$-preferred repair of $\mathcal{DB}$. Then there is a $\leq_i$-maximally consistent element $N$ in $\mathcal{M}^{\mathcal{DB}}$ s.t.* $\mathsf{Insert} = \mathsf{Insert}^N$ *and* $\mathsf{Retract} = \mathsf{Retract}^N$.

*Proof:* The pair $(\mathsf{Insert},\ \mathsf{Retract})$ is in particular a repair of $\mathcal{DB}$, thus by Proposition 2 there is a classical model $M$ of $\mathcal{IC}$ such that $\mathsf{Insert} = M^t \setminus \mathcal{D}$ and $\mathsf{Retract} = \mathcal{D} \setminus M^t$. Consider the following valuation:

$$N(p) = \begin{cases} \top & \text{if } p \in M^t \setminus \mathcal{D} \text{ or } p \in \mathcal{D} \setminus M^t \\ M(p) & \text{otherwise.} \end{cases}$$

First we show that $N = \mathcal{H}^{\mathcal{D}} \oplus M$. This is so since if $M(p) = \mathcal{H}^{\mathcal{D}}(p)$, then since $\mathcal{H}^{\mathcal{D}}$ is a minimal Herbrand model of $\mathcal{D}$, necessarily $p \notin M^t \setminus \mathcal{D}$ and $p \notin \mathcal{D} \setminus M^t$, thus $N(p) = M(p) = M(p) \oplus M(p) = M(p) \oplus \mathcal{H}^{\mathcal{D}}(p)$. Otherwise, if $M(p) \neq \mathcal{H}^{\mathcal{D}}(p)$, then either $M(p) = t$ and $\mathcal{H}^{\mathcal{D}}(p) = f$, i.e., $p \in M^t \setminus \mathcal{D}$, or $M(p) = f$ and $\mathcal{H}^{\mathcal{D}}(p) = t$, i.e., $p \in \mathcal{D} \setminus M^t$. In both cases, $N(p) = \top = M(p) \oplus \mathcal{H}^{\mathcal{D}}(p)$[17]. Thus $N = \mathcal{H}^{\mathcal{D}} \oplus M$, and so $N \in \mathcal{M}^{\mathcal{DB}}$. Now, by Proposition 2 again, and by the definition of $N$, $\mathsf{Insert}^N = N^{\top} \setminus \mathcal{D} = [(M^t \setminus \mathcal{D}) \cup (\mathcal{D} \setminus M^t)] \setminus \mathcal{D} = M^t \setminus \mathcal{D} = \mathsf{Insert}$, and $\mathsf{Retract}^N = N^{\top} \cap \mathcal{D} = [(M^t \setminus \mathcal{D}) \cup (\mathcal{D} \setminus M^t)] \cap \mathcal{D} = \mathcal{D} \setminus M^t = \mathsf{Retract}$.

It remains to show that $N$ is $\leq_i$-maximally consistent in $\mathcal{M}^{\mathcal{DB}}$. Suppose not. Then there is an $N' \in \mathcal{M}^{\mathcal{DB}}$ s.t. $(N')^{\top} \subset N^{\top} = \mathsf{Insert} \cup \mathsf{Retract}$. By Proposition 3, $(\mathsf{Insert}^{N'},\ \mathsf{Retract}^{N'})$ is also a repair of $\mathcal{DB}$. Moreover,
- $\mathsf{Insert}^{N'} = (N')^{\top} \setminus \mathcal{D} \subseteq N^{\top} \setminus \mathcal{D} = \mathsf{Insert}^N = \mathsf{Insert}$,
- $\mathsf{Retract}^{N'} = (N')^{\top} \cap \mathcal{D} \subseteq N^{\top} \cap \mathcal{D} = \mathsf{Retract}^N = \mathsf{Retract}$,
- $\mathsf{Insert}^{N'} \cup \mathsf{Retract}^{N'} = (N')^{\top} \subset N^{\top} = \mathsf{Insert}^N \cup \mathsf{Retract}^N = \mathsf{Insert} \cup \mathsf{Retract}$.

Hence $(\mathsf{Insert}^{N'},\ \mathsf{Retract}^{N'}) <_i (\mathsf{Insert},\ \mathsf{Retract})$, and so $(\mathsf{Insert},\ \mathsf{Retract})$ is not a $\leq_i$-preferred repair of $(\mathcal{D}, \mathcal{IC})$, a contradiction. □

Propositions 5 and 6 may be formulated in terms of $\leq_c$ as follows:

17. Here we use the fact that $t \oplus f = \top$.

**Proposition 7** *If $N$ is a $\leq_c$-maximally consistent element in $\mathcal{M}^{\mathcal{DB}}$, then* $(\mathsf{Insert}^N,\ \mathsf{Retract}^N)$ *is a $\leq_c$-preferred repair of $\mathcal{DB}$.*

**Proposition 8** *Suppose that* $(\mathsf{Insert},\ \mathsf{Retract})$ *is a $\leq_c$-preferred repair of $\mathcal{DB}$. Then there is a $\leq_c$-maximally consistent element $N$ in $\mathcal{M}^{\mathcal{DB}}$ s.t.* $\mathsf{Insert} = \mathsf{Insert}^N$ *and* $\mathsf{Retract} = \mathsf{Retract}^N$.

The proofs of the last two propositions are similar to those of Propositions 5 and 6, respectively.

**Example 6** Consider again Example 3. We have that:

$$\mathcal{UDB} = (\mathcal{D}, \mathcal{IC}) = (\,\{p(a),\ p(b),\ q(a),\ q(c)\},\ \{\forall X(p(X)\rightarrow q(X))\}\,).$$

Thus, $\mathcal{H}^{\mathcal{D}} = \{p(a):t,\ p(b):t,\ p(c):f,\ q(a):t,\ q(b):f,\ q(c):t\}$, and the classical models of $\mathcal{IC}$ are those in which either $p(y)$ is false or $q(y)$ is true for every $y\in\{a,b,c\}$. Now, since in $\mathcal{H}^{\mathcal{D}}$ neither $p(b)$ is false nor $q(b)$ is true, it follows that *every* element in $\mathcal{M}^{\mathcal{UDB}}$ must assign $\top$ either to $p(b)$ or to $q(b)$. Hence, the $\leq_i$-maximally consistent elements in $\mathcal{M}^{\mathcal{UDB}}$ (which in this case are also the $\leq_c$-maximally consistent elements in $\mathcal{M}^{\mathcal{UDB}}$) are the following:

$$M_1 = \{\, p(a){:}t,\ p(b){:}\top,\ p(c){:}f,\ q(a){:}t,\ \ q(b){:}f,\ q(c){:}t\,\},$$
$$M_2 = \{\, p(a){:}t,\ p(b){:}t,\ p(c){:}f,\ q(a){:}t,\ q(b){:}\top,\ q(c){:}t\,\}.$$

By Propositions 5 and 6, then, the $\leq_i$-preferred repairs of $\mathcal{UDB}$ (which are also its $\leq_c$-preferred repairs) are $(\mathsf{Insert}^{M_1},\ \mathsf{Retract}^{M_1}) = (\emptyset,\ \{p(b)\})$ and $(\mathsf{Insert}^{M_2},\ \mathsf{Retract}^{M_2}) = (\{q(b)\},\ \emptyset)$ (cf. Example 3).

**Example 7** In Examples 4 and 5, the $\leq_i$-maximally consistent elements (and the $\leq_c$-maximally consistent elements) of $\mathcal{M}^{\mathcal{DB}}$ are $N_1 = \{p{:}t, q{:}\top, r{:}t\}$ and $N_2 = \{p{:}\top, q{:}f, r{:}t\}$. It follows that the preferred repairs in this case are $(\{q\},\ \emptyset)$ and $(\emptyset,\ \{p\})$.

To summarize, in this section we have considered a model-based, three-valued *preferential semantics* for database integration. We have shown (Propositions 5 – 8) that common and natural criteria for making preferences among possible repairs (i.e., set inclusion and minimal cardinality) can be expressed by order relations on three-valued models of the database. The two ways of making preferences (among repairs on one hand and among three-valued models on the other hand) are thus strongly related, and induce two alternative approaches for database integration. In the next section we shall consider a third approach to the same problem (aimed to provide an *operational semantics* for database integration) and relate it to the model-based semantics, discussed above.

## 4. Computing Repairs through Abduction

In this section we introduce an abductive system that consistently integrates possibly contradicting data-sources. This system computes, for a set of data-sources and a preference criterion $\leq$, the corresponding $\leq$-repaired databases[18]. Our framework is composed of an abductive logic program (Denecker & Kakas, 2000) and an abductive solver $\mathcal{A}$system (Kakas, Van Nuffelen, & Denecker, 2001; Van Nuffelen & Kakas, 2001) that is based on the

18. It is important to note already in this stage that for computing the $\leq$-repaired databases it won't be necessary to produce all the repaired databases.

abductive refutation procedure SLDNFA (Denecker & De Schreye, 1992, 1998). In the first three parts of this section we describe these components: in Section 4.1 we give a general description of abductive reasoning, in Section 4.2 we show how it can be applied to encode database repairs, and in Section 4.3 we describe the 'computational platform'. Then, in Section 4.4 we demonstrate the computation process by a comprehensive example, and in Section 4.5 we specify soundness and completeness results of our approach (with respect to the basic definitions of Section 2 and the model-based semantics of Section 3). Finally, in Section 4.6 we consider some ways of representing special types of data in the system.

### 4.1 Abductive Logic Programming

We start with a general description of abductive reasoning in the context of logic programming. As usual in logic programming, the language contains constants, functions, and predicate symbols. A *term* is either a variable, a constant, or a compound term $f(t_1, \ldots, t_n)$, where $f$ is an $n$-ary function symbol and $t_i$ are terms. An *atom* is an expression of the form $p(t_1, \ldots, t_m)$, where $p$ is an $m$-ary predicate symbol and $t_i$ ($i = 1, \ldots, m$) are terms. A *literal* is an atom or a negated atom. A *denial* is an expression of the form $\forall \overline{X}(\leftarrow F)$, where $F$ is a conjunction of literals and $\overline{X}$ is a subset of the variables in $F$. The free variables in $F$ (those that are not in $\overline{X}$) can be considered as place holders for objects of unspecified identity (Skolem constants). Intuitively, the body $F$ of a denial $\forall \overline{X}(\leftarrow F)$ represents an invalid situation.

**Definition 13** (Kakas et al., 1992; Denecker & Kakas, 2000) An *abductive logic theory* is a triple $\mathcal{T} = (\mathcal{P}, \mathcal{A}, \mathcal{IC})$, where:

- $\mathcal{P}$ is a *logic program*, consisting of clauses of the form $h \leftarrow l_1 \wedge \ldots \wedge l_n$, where $h$ is an atomic formula and $l_i$ ($i = 1, \ldots, n$) are literals. These clauses are interpreted as definitions for the predicates in their heads,
- $\mathcal{A}$ is a set of *abducible predicates*, i.e., predicates that do not appear in the head of any clause in $\mathcal{P}$,
- $\mathcal{IC}$ is a set of first-order formulae, called *integrity constraints*.

All the main model semantics of logic programming can be extended to abductive logic programming. This includes two-valued completion (Console, Theseider Dupre, & Torasso, 1991) and three-valued completion semantics (Denecker & De Schreye, 1993), extended well-founded semantics (Pereira, Aparicio, & Alferes, 1991), and generalized stable semantics (Kakas & Mancarella, 1990b). These semantics can be defined in terms of arbitrary interpretations (Denecker & De Schreye, 1993), but generally they are based on Herbrand interpretations. The effect of this restriction on the semantics of the abductive theory is that a *domain closure* condition is imposed: the domain of interpretation is known to be the Herbrand universe. A *model* of an abductive theory under any of these semantics is a Herbrand interpretation $\mathcal{H}$, for which there exists a collection of ground abducible facts $\Delta$, such that $\mathcal{H}$ is a model of the logic program $\mathcal{P} \cup \Delta$ (with respect to the corresponding semantics of logic programming) and $\mathcal{H}$ classically satisfies any element in $\mathcal{IC}$.

Similarly, for any of the main semantics $\mathcal{S}$ of logic programming, one can define the notion of an abductive solution for a query and an abductive theory.

**Definition 14** (Kakas et al., 1992; Denecker & Kakas, 2000) An (abductive) *solution* for a theory $(\mathcal{P}, \mathcal{A}, \mathcal{IC})$ and a query $\mathcal{Q}$ is a set $\Delta$ of ground abducible atoms, each one having a predicate symbol in $\mathcal{A}$, together with an answer substitution $\theta$, such that the following three conditions are satisfied:

a) $\mathcal{P} \cup \Delta$ is consistent in the semantics $\mathcal{S}$,

b) $\mathcal{P} \cup \Delta \models_{\mathcal{S}} \mathcal{IC}$,

c) $\mathcal{P} \cup \Delta \models_{\mathcal{S}} \forall \mathcal{Q}\theta$.

In the next section we will use an abductive theory with a non-recursive program to model the database repairs. The next proposition shows that for such abductive theories all Herbrand semantics coincide, and models correspond to abductive solutions for the query `true`.

**Proposition 9** *Let $\mathcal{T} = (\mathcal{P}, \mathcal{A}, \mathcal{IC})$ be an abductive theory, such that $\mathcal{P}$ is a non-recursive program. Then $\mathcal{H}$ is a Herbrand model of the three-valued completion semantics, iff $\mathcal{H}$ is a Herbrand model of the two-valued completion semantics, iff $\mathcal{H}$ is a generalized stable model of $\mathcal{T}$, iff $\mathcal{H}$ is a generalized well-founded model of $\mathcal{T}$.*

*If $\mathcal{H}$ is a model of $\mathcal{T}$, then the set $\Delta$ of abducible atoms in $\mathcal{H}$ is an abductive solution for the query* `true`. *Conversely, for every abductive solution for* `true`, *there exists a unique model $\mathcal{H}$ of $\mathcal{T}$, such that $\Delta$ is the set of true abducible atoms in $\mathcal{H}$.*

*Proof:* The proof is based on the well-known fact that for non-recursive logic programs, all the main semantics of logic programming coincide. In particular, for a non-recursive logic program $\mathcal{P}$ there is a Herbrand interpretation $\mathcal{H}$, which is the unique model under each semantics (see, for example, Denecker & De Schreye, 1993).

Let $\mathcal{H}$ be a model of $\mathcal{T} = (\mathcal{P}, \mathcal{A}, \mathcal{IC})$ under any of the four semantics mentioned above. Then there exists a collection of ground abducible facts $\Delta$, such that $\mathcal{H}$ is a model of the logic program $\mathcal{P} \cup \Delta$ under the corresponding semantics of logic programming. Since $\mathcal{P}$ is non-recursive, so is $\mathcal{P} \cup \Delta$. By the above observation, $\mathcal{H}$ is the unique model of $\mathcal{P} \cup \Delta$ under any of the above mentioned semantics. Hence, $\mathcal{H}$ is a model of $\mathcal{T}$ under any of the other semantics. This proves the first part of the proposition.

When $\mathcal{H}$ is a Herbrand model of $\mathcal{T}$, there is a set $\Delta$ of abducible atoms such that $\mathcal{H}$ is a model of $\mathcal{P} \cup \Delta$. Clearly, $\Delta$ must be the set of true abducible atoms in $\mathcal{H}$. Then $\mathcal{P} \cup \Delta$ is obviously consistent, and it entails the integrity constraints of $\mathcal{T}$, which entails `true`. Hence, $\Delta$ is an abductive solution for `true`. Conversely, for any set $\Delta$ of abducible atoms, $P \cup \Delta$ has a unique model $\mathcal{H}$ and the set of true abducible atoms in $\mathcal{H}$ is $\Delta$. When $\Delta$ is an abductive solution for `true`, $\mathcal{H}$ satisfies the integrity constraints, and hence $\mathcal{H}$ is a model of $\mathcal{T}$. Consequently, $\mathcal{H}$ is the unique model of $\mathcal{T}$, and its set of true abducible atoms is $\Delta$. □

In addition to the standard properties of abductive solutions for a theory $\mathcal{T}$ and a query $\mathcal{Q}$, specified in Definition 14, one frequently imposes optimization conditions on the solutions $\Delta$, analogous to those found in the context of database repairs. Two frequently used criteria are that the generated abductive solution $\Delta$ should be minimal with respect to set inclusion or with respect to set cardinality (cf. Definition 7). The fact that the same preference criteria are used for choosing appropriate abductive solutions and for selecting preferred database repairs does not necessarily mean that there is a natural mapping between the corresponding solutions. In the next sections we will show, however, that *meta-programming* allows us to map a database repair problem into an abductive problem (w.r.t. the same type of preference criterion).

### 4.2 An Abductive Meta-Program for Encoding Database Repairs

The task of repairing the union of $n$ given databases $\mathcal{DB}_i$ with respect to the integration of the local integrity constraints $\mathcal{IC}$, can be represented by an abductive theory $\mathcal{T} = (\mathcal{P}\ , \mathcal{A}\ , \mathcal{IC}')$, where $\mathcal{P}$ is a meta-program encoding how a new database is obtained by updating the existing databases, $\mathcal{A}$ is the set $\{\texttt{insert}, \texttt{retract}\}$ of abducible predicates used to describe updates, and $\mathcal{IC}'$ encodes the integrity constraints. In $\mathcal{P}$, facts $p$ that appear in at least one of the databases are encoded by atomic rules $\texttt{db}(p)$, and facts $p$ that appear in the updated database are represented by atoms $\texttt{fact}(p)$. The latter predicate is defined as follows:

```
fact(X) ← db(X) ∧ ¬retract(X)
fact(X) ← insert(X)
```

To assure that the predicates `insert` and `retract` encode a proper update of the database, the following integrity constraints are also specified:

- An inserted element should not belong to a given database:
  `← insert(X) ∧ db(X)`
- A retracted element should belong to some database:
  `← retract(X) ∧ ¬db(X)`

The set of integrity constraints $\mathcal{IC}'$ is obtained by a straightforward transformation from $\mathcal{IC}$: every occurrence of a database fact $p$ in some integrity constraint is replaced by $\texttt{fact}(p)$[19].

**Example 8 (Example 1, revisited)** Figure 2 contains the meta-program encoding Example 1 (the codes for Examples 2 and 3 are similar).

As noted in Section 4.1, under any of the main semantics of abductive logic programing there is a one to one correspondence between repairs of the composed database $\mathcal{DB}$ and the Herbrand models of its encoding, the abductive meta theory $\mathcal{T}$. Consequently, abduction can be used to compute repairs. In the following sections we introduce an abductive method for this purpose.

19. Since our abductive system (see Section 4.3) will accept integrity constraints in a denial form, in case that the elements of $\mathcal{IC}'$ are not in this form, the Lloyd-Topor transformation (Lloyd & Topor, 1984) may also be applied here; we consider this case in Section 4.3.2.

```
% System definitions:
defined(fact(_))
defined(db(_))
abducible(insert(_))
abducible(retract(_))

% The composer:
fact(X) ← db(X) ∧ ¬retract(X)
fact(X) ← insert(X)
← insert(X) ∧ db(X)
← retract(X) ∧ ¬db(X)

% The databases:
db(teaches(c1,n1))                                % D1
db(teaches(c2,n2))
db(teaches(c2,n3))                                % D2
Y = Z ← fact(teaches(X,Y)) ∧ fact(teaches(X,Z))   % IC
```

Figure 2: A meta-program for Example 1

### 4.3 The Abductive Computational Model (The $\mathcal{A}$system)

Below we describe the abductive system that will be used to compute database repairs. The $\mathcal{A}$system (Kakas, Van Nuffelen, & Denecker, 2001; Van Nuffelen & Kakas, 2001) is a tool combining abductive logic theories and constraint logic programming (CLP). It is a synthesis of the refutation procedures SLDNFA (Denecker & De Schreye, 1998) and ACLP (Kakas et al., 2000), together with an improved control strategy. The essence of the $\mathcal{A}$system is a reduction of a high level specification to a lower level constraint store, which is managed by a constraint solver. See `http://www.cs.kuleuven.ac.be/~dtai/kt/` for the latest version of the system[20]. Below we review the theoretical background as well as some practical considerations behind this system. For more information, see (Denecker & De Schreye, 1998) and (Kakas, Van Nuffelen, & Denecker, 2001).

#### 4.3.1 Abductive Inference

The input to the $\mathcal{A}$system is an abductive theory $\mathcal{T} = (\mathcal{P}, \mathcal{A}, \mathcal{IC})$, where $\mathcal{IC}$ consists of universally quantified denials. The process of answering a query $\mathcal{Q}$, given by a conjunction of literals, can be described as a derivation for $\mathcal{Q}$ through rewriting *states.* A state is a pair $(\mathcal{G}\ ,\ ST)$, where $\mathcal{G}$, the set of *goal formulae,* is a set of conjunctions of literals and denials. During the rewriting process the elements in $\mathcal{G}$ (the goals) are reduced to basic formulae

20. This version runs on top of Sicstus Prolog 3.10.1 or later versions.

that are stored in the structure $ST$. This structure is called a *store*, and it consists of the following elements[21]:

- a set $\Delta$ that contains abducibles $a(\overline{t})$.
- a set $\Delta^*$ that contains denials of the form $\forall \overline{X}(\leftarrow a(\overline{t}) \wedge Q)$, where $a(\overline{t})$ is an abducible. Such a denial may contain free variables.
- a set $\mathcal{E}$ of equalities and inequalities over terms.

The consistency of $\mathcal{E}$ is maintained by a constraint solver that uses the Martelli and Montanari unification algorithm (Martelli & Montanari, 1982) for the equalities and constructive negation for the inequalities.

A state $\mathcal{S} = (\mathcal{G}\,,\, ST)$ is called *consistent* if $\mathcal{G}$ does not contain `false` and $ST$ is consistent (since $\Delta$ and $\Delta^*$ are kept consistent with each other and with $\mathcal{E}$, the latter condition is equivalent to the consistency of $\mathcal{E}$). A consistent state with an empty set of goals ($\mathcal{G} = \emptyset$) is called a *solution state*.

A derivation starts with an *initial state* $(\mathcal{G}_0\,,\, ST_0)$, where every element in $ST_0$ is empty, and the initial goal, $\mathcal{G}_0$, contains the query $\mathcal{Q}$ and all the integrity constraints $\mathcal{IC}$ of the theory $\mathcal{T}$. Then a sequence of rewriting *steps* is performed. A step starts in a certain state $\mathcal{S}_i = (\mathcal{G}_i\,,\, ST_i)$, selects a goal in $\mathcal{G}_i$, and applies an inference rule (see below) to obtain a new consistent state $\mathcal{S}_{i+1}$. When no consistent state can be reached from $\mathcal{S}_i$ the derivation backtracks. A derivation terminates when a solution state is reached, otherwise it fails (see Section 4.4 below for a demonstration of this process).

Next we present the inference rules in the $\mathcal{A}$system, using the following conventions:

- $\mathcal{G}_i^- = \mathcal{G}_i - \{F\}$, where $F$ is the selected goal formula.
- **OR** and **SELECT** denote nondeterministic choices in an inference rule.
- Q is a conjunction of literals, possibly empty. Since an empty conjunction is equivalent to `true`, the denial $\leftarrow Q$ with empty $Q$ is equivalent to `false`.
- If $\Delta$, $\Delta^*$, and $\mathcal{E}$ are not mentioned, they remain unchanged.

The inference rules are classified in four groups, named after the leftmost literal in the selected formula (shown in bold). Each group contains rules for (positive) conjunctions of literals and rules for denials.

1. *Defined predicates:*
   The inference rules unfold the bodies of a defined predicate. For positive conjunctions this corresponds to standard resolution with a selected clause, whereas in the denial all clauses are used because every clause leads to a new denial.

   D.1 $\mathbf{p(\overline{t}) \wedge Q}$:
   Let $p(\overline{s}_i) \leftarrow B_i \in \mathcal{P}$ $(i = 1, \ldots, n)$ be $n$ clauses with $p$ in the head. Then:
   $\mathcal{G}_{i+1} = \mathcal{G}_i^- \cup \{\overline{t} = \overline{s}_1 \wedge B_1 \wedge Q\}$ **OR** ... **OR** $\mathcal{G}_{i+1} = \mathcal{G}_i^- \cup \{\overline{t} = \overline{s}_n \wedge B_n \wedge Q\}$

21. The actual implementation of the $\mathcal{A}$system also contains a store for finite domain constraint expressions. This store is not needed for the application here, and hence it is omitted.

D.2 $\forall \overline{\mathbf{X}}(\leftarrow \mathbf{p}(\overline{\mathbf{t}}) \wedge \mathbf{Q})$:
$\mathcal{G}_{i+1} = \mathcal{G}_i^- \cup \{\forall \overline{X}, \overline{Y}(\leftarrow \overline{t} = \overline{s} \wedge B \wedge Q) \,|\, \text{there is } p(\overline{s}) \leftarrow B \in \mathcal{P} \text{ with variables } \overline{Y}\}$

2. *Negations:*
Resolving negation corresponds to 'switching' the mode of reasoning from a positive literal to a denial and vice versa. This is similar to the idea of negation-as-failure in logic programming.

N.1 $\neg \mathbf{p}(\overline{\mathbf{t}}) \wedge \mathbf{Q}$:
$\mathcal{G}_{i+1} = \mathcal{G}_i^- \cup \{Q, \leftarrow p(\overline{t})\}$

N.2 $\forall \overline{\mathbf{X}}(\leftarrow \neg \mathbf{p}(\overline{\mathbf{t}}) \wedge \mathbf{Q})$ and $\overline{t}$ does not contain variables in $\overline{X}$:
$\mathcal{G}_{i+1} = \mathcal{G}_i^- \cup \{p(\overline{t})\}$ **OR** $\mathcal{G}_{i+1} = \mathcal{G}_i^- \cup \{\leftarrow p(\overline{t}), \forall \overline{X}(\leftarrow Q)\}$

3. *Abducibles:*
The first rule is responsible for the creation of new hypotheses. Both rules ensure that the elements in $\Delta$ are consistent with those in $\Delta^*$.

A.1 $\mathbf{a}(\overline{\mathbf{t}}) \wedge \mathbf{Q}$:
**SELECT** an arbitrary $a(\overline{s}) \in \Delta$ and define $\mathcal{G}_{i+1} = \mathcal{G}_i^- \cup \{Q\} \cup \{\overline{s} = \overline{t}\}$
**OR** $\mathcal{G}_{i+1} = \mathcal{G}_i^- \cup \{Q\} \cup \{\forall \overline{X}(\leftarrow \overline{s} = \overline{t} \wedge R) \,|\, \forall \overline{X}(\leftarrow a(\overline{s}) \wedge R) \in \Delta_i^*\} \cup$
$\{\leftarrow (\overline{t} = \overline{s}) \,|\, a(\overline{s}) \in \Delta_i\}$ and $\Delta_{i+1} = \Delta_i \cup \{a(\overline{t})\}$

A.2 $\forall \overline{\mathbf{X}}(\leftarrow \mathbf{a}(\overline{\mathbf{t}}) \wedge \mathbf{Q})$:
$\mathcal{G}_{i+1} = \mathcal{G}_i^- \cup \{\forall \overline{X}(\leftarrow \overline{s} = \overline{t} \wedge Q) \,|\, a(\overline{s}) \in \Delta_i\}$ and $\Delta_{i+1}^* = \Delta_i^* \cup \{\forall \overline{X}(\leftarrow a(\overline{t}) \wedge Q)\}$

4. *Equalities:*
These inference rules isolate the (in)equalities, so that the constraint solver can evaluate them. The first rule applies to equalities in goal formulae:

E.1 $\mathbf{s} = \mathbf{t} \wedge \mathbf{Q}$:
$\mathcal{G}_{i+1} = \mathcal{G}_i^- \cup \{Q\}$ and $\mathcal{E}_{i+1} = \mathcal{E}_i \cup \{s = t\}$

The following three rules handle equalities in denials. Which rule applies depends on whether $s$ or $t$ contain free or universally quantified variables. In these rules $Q[X/t]$ denotes the formula that is obtained from $Q$ by substituting the term $t$ for $X$.

E.2 $\forall \overline{\mathbf{X}}(\leftarrow \mathbf{s} = \mathbf{t} \wedge \mathbf{Q})$:
If $s$ and $t$ are not unifiable then $\mathcal{G}_{i+1} = \mathcal{G}_i^-$;
Otherwise, let $E_s$ be the equation set in solved form representing a most general unifier of $s$ and $t$ (Martelli & Montanari, 1982). $\mathcal{G}_{i+1} = \mathcal{G}_i^- \cup \{\forall \overline{X}(\leftarrow E_s \wedge Q)\}$.

E.3 $\forall \mathbf{X}, \overline{\mathbf{Y}}(\leftarrow \mathbf{X} = \mathbf{t} \wedge \mathbf{Q})$ where $t$ is a term not containing $X$:
$\mathcal{G}_{i+1} = \mathcal{G}_i^- \cup \{\forall \overline{Y}(\leftarrow Q[X/t])\}$

E.4 $\forall \overline{\mathbf{X}}, \overline{\mathbf{Y}}(\leftarrow \mathbf{X} = \mathbf{t} \wedge \mathbf{Q})$ where $X$ is a free variable and $\overline{X}$ is the set of universally quantified variables in a term $t$:
$\mathcal{E}_{i+1} = \mathcal{E}_i \cup \{\forall \overline{X}(X \neq t)\}$ **OR** $\mathcal{G}_{i+1} = \mathcal{G}_i^- \cup \{X = t\} \cup \{\forall \overline{Y}(\leftarrow Q[X/t])\}$[22].

22. In its first branch the inference E.4 explores the condition $\forall \overline{X}(X \neq t)$. In the second branch, the negation of this condition is explored. Here $X$ is identical to $t$, for some values assigned to $\overline{X}$. This is why in the second branch, the universally quantified variables $\overline{X}$ are turned into free variables which may appear free in $\forall \overline{Y}(\leftarrow Q[X/t])$.

As usual, one has to check for *floundering negation*. This occurs when the inference rule N.2 is applied on a denial with universally quantified variables in the negative literal $\neg p(\bar{t})$. Floundering aborts the derivation.

An answer substitution $\theta$, derived from a solution state $\mathcal{S}$, is any substitution $\theta$ of the free variables in $\mathcal{S}$ which satisfies $\mathcal{E}$ (i.e. $\theta(\mathcal{E})$ is true) and grounds $\Delta$. Note that, in case of an abductive theory without abducibles and integrity constraints, computed answers as defined by Lloyd (1987) are most general unifiers of $\mathcal{E}$ and correct answers are answer substitutions as defined above.

**Proposition 10** (Kakas, Van Nuffelen, & Denecker, 2001) *Let $\mathcal{T} = (\mathcal{P}, \mathcal{A}, \mathcal{IC})$ be an abductive theory, $\mathcal{Q}$ a query, $S$ a solution state of a derivation for $\mathcal{Q}$, and $\theta$ an answer substitution of $S$. Then the pair consisting of the ground abducible atoms $\theta(\Delta(\mathcal{S}))$ and of the answer substitution $\theta$ is an abductive solution for $\mathcal{T}$ and $\mathcal{Q}$.*

#### 4.3.2 Constraint Transformation to Denial Form

Since the inference rules of the $\mathcal{A}$system are applied only on integrity constraints in denial form, the integrity constraints $\mathcal{IC}$ in the abductive theory $\mathcal{T}$ must be translated to this form. This is done by applying a variant of the Lloyd-Topor transformation (Lloyd & Topor, 1984) on the integrity constraints (see Denecker & De Schreye, 1998). This is the same procedure as the well-known procedure used in deductive databases to convert a first order quantified query $\mathcal{Q}$ into a logically equivalent pair of an atomic query and a non-recursive datalog procedure. The transformation is defined as a rewriting process of sets of formulae: the initial set is $\{\leftarrow \neg F | F \in \mathcal{IC}\}$, and the transformation is done by applying De Morgan and various distribution rules. New predicates and rules may be introduced during the transformation in order to deal with universal quantifications in denials. Below we illustrate the transformation in the case of the integrity constraints of the running example.

**Example 9** Consider the following extension of the integrity constraints of Example 1:

$$\mathcal{IC} = \{\, \forall X \forall Y \forall Z\ (teaches(X,Y) \wedge teaches(X,Z)\ \rightarrow\ Y = Z)\,,$$
$$\forall X\ (teacher(X)\ \rightarrow\ \exists Y\, teaches(Y,X))\ \}.$$

Note that in addition to the original integrity constraint of Example 1, here we also demand that every teacher has to give at least one course.

- Lloyd-Topor transformation on the first integrity constraint:

  **(1)** $\leftarrow \neg \forall X\, \forall Y\, \forall Z\ (\neg teaches(X,Y) \vee \neg teaches(X,Z) \vee Y = Z)$

  **(2)** $\leftarrow \exists X\, \exists Y\, \exists Z\ (teaches(X,Y) \wedge teaches(X,Z) \wedge Y \neq Z)$

  **(3)** $\forall X\, \forall Y\, \forall Z\ (\leftarrow teaches(X,Y) \wedge teaches(X,Z) \wedge Y \neq Z)$

- Lloyd-Topor transformation on the second integrity constraint:

  **(1)** $\leftarrow \neg \forall X\ (\neg teacher(X) \vee \exists Y\, teaches(Y,X))$

  **(2)** $\forall X\ (\leftarrow teacher(X) \wedge \neg \exists Y\, teaches(Y,X))$

**(3)** $\forall X\ (\leftarrow teacher(X) \wedge \neg gives_courses(X))$ where $gives_courses$ is defined by:
$gives_courses(X) \leftarrow \exists Y\ teaches(Y, X)$

**(4)** $\forall X\ (\leftarrow teacher(X) \wedge \neg gives_courses(X))$, and
$gives_courses(X) \leftarrow teaches(Y, X)$

### 4.3.3 Control Strategy

The selection strategy applied during the derivation process is crucial. A Prolog-like selection strategy (left first, depth first) often leads to trashing, because it is blind to other choices, and it does not result in a global overview of the current state of the computation. In the development of the $\mathcal{A}$system the main focus was on the improvement of the control strategy. The idea is to apply first those rules that result in a deterministic change of the state, so that information is propagated. If none of such rules is applicable, then one of the left over choices is selected. By this strategy, commitment to a choice is suspended until the moment where no other information can be derived in a deterministic way. This resembles a CLP-solver, in which the constraints propagate their information as soon as a choice is made. This propagation can reduce the number of choices to be made and thus often dramatically increases the performance.

### 4.3.4 Implementation

In this section we describe the structure of our implementation. Figure 3 shows a layered view. The upper-most level consists of the specific abductive logic theory of the integration task, i.e., the database information and the integrity constraints. This layer together with the composer form the abductive meta-theory (see Section 4.2) that is processed by the $\mathcal{A}$system.

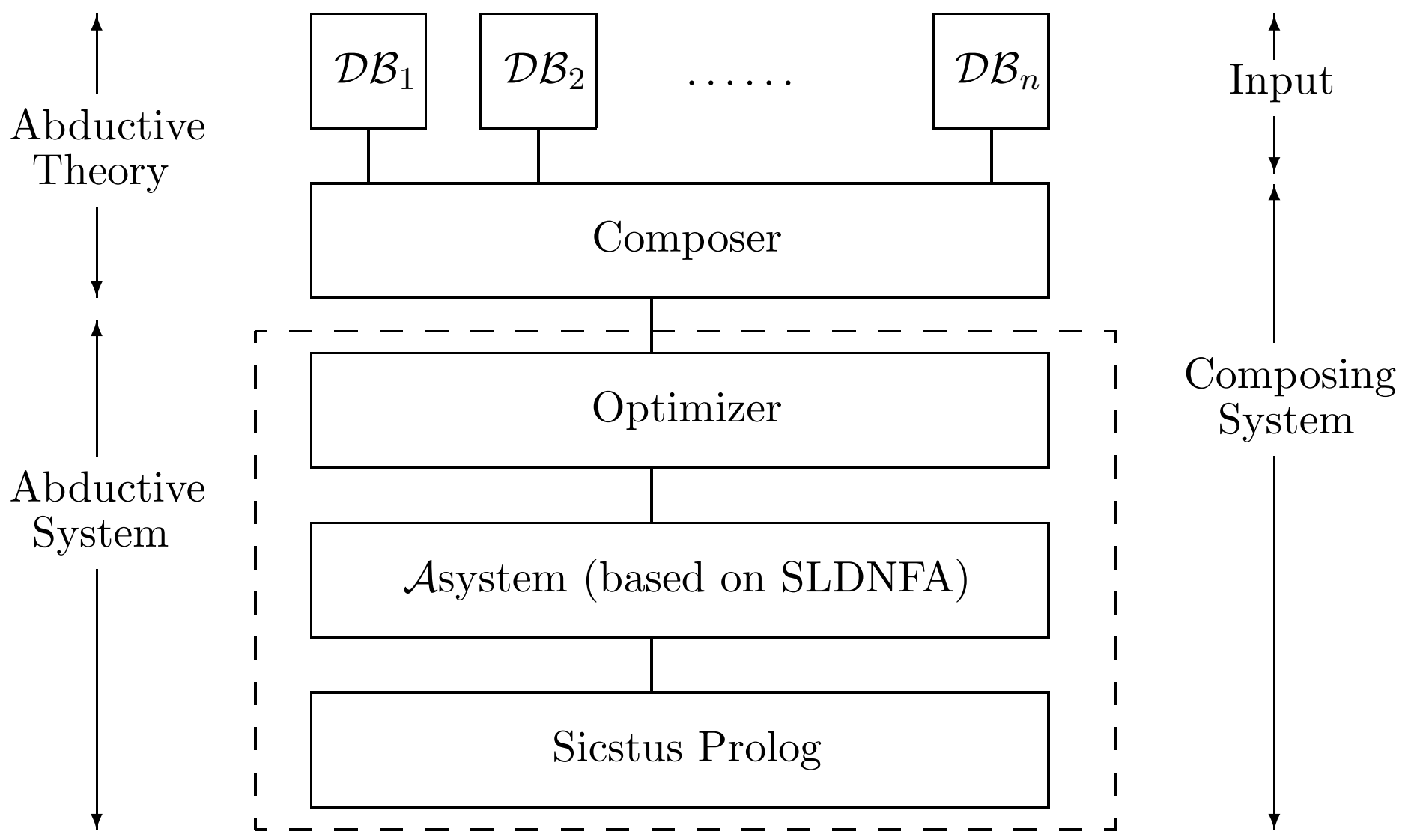

Figure 3: A schematic view of the system components.

As noted above, the *composer* consists of a meta-theory for integrating the databases in a coherent way. It is interpreted here as an abductive theory, in which the abducible predicates provide the information on how to restore the consistency of the amalgamated data.

The abductive system (enclosed by dotted lines in Figure 3) consists of three main components: a finite domain constraint solver (part of Sicstus Prolog), an abductive meta-interpreter (described in the previous sections), and an optimizer.

The *optimizer* is a component that, given a preference criterion on the space of the solutions, computes only the most-preferred (abductive) solutions. Given such a preference criterion, this component prunes 'on the fly' those branches of the search tree that lead to solutions that are worse than others that have already been computed. This is actually a branch and bound 'filter' on the solutions space, that speeds-up execution and makes sure that only the desired solutions will be obtained[23]. If the preference criterion is a pre-order, then the optimizer is *complete*, that is, it can compute all the optimal solutions (more about this in Section 4.5). Moreover, this is a general-purpose component, and it may be useful not only for data integration, but also for, e.g., solving planning problems.

#### 4.3.5 Complexity

It is well-known that in general, the task of repairing a database is not tractable, as there may be an exponential number of different ways of repairing it. Even in cases where integrity constraints are assumed to be single-headed dependencies (Greco & Zumpano, 2000), checking whether there exists a $\leq$-repaired database in which a certain query $\mathcal{Q}$ is satisfied, is in $\Sigma_2^P$. Checking if a fact is satisfied by *all* the $\leq$-repaired databases is in $\Pi_2^P$ (see Greco & Zumpano, 2000). This is not surprising in light of the correspondence between computations of $\leq$-minimal repairs and computations of entailment relations defined by maximally consistent models (see Propositions 5–8), also known to be on the second level of the polynomial hierarchy.

A pure upper bound for the $\mathcal{A}$system is still unknown, since – to the best of our knowledge – no complexity results on SLDNFA refutation procedure are available.

### 4.4 Example: A Derivation of Repairs by the $\mathcal{A}$system

Consider again Example 9. The corresponding meta-theory (assuming that the Lloyd-Topor transformation has been applied on it) is given in Figure 4. In this case, and in what follows, we shall assume that all variables in the denials are universally quantified, and so, in order to reduce the amount of notations, universal quantifiers are omitted from the denial rules.

We have executed the code of Figure 4, as well as other examples from the literature in our system. As Theorem 2 in Section 4.5 guarantees, the output in each case is the set of the most preferred solutions of the corresponding problem. In what follows we demonstrate

23. See also the third item of Note 2 (at the end of Section 4.4).

```
db(teacher(n1))
db(teacher(n2))
db(teacher(n3))
db(teaches(c1,n1))
db(teaches(c2,n2))
db(teaches(c2,n3))

← fact(teaches(X,Y)) ∧ fact(teaches(X,Z)) ∧ (Y ≠ Z)   (ic1)
← fact(teacher(X)) ∧ ¬gives_courses(X)                (ic2)
gives_courses(X) ← fact(teaches(Y,X))

fact(X) ← db(X) ∧ ¬retract(X)
fact(X) ← insert(X)
← insert(X) ∧ db(X)                          (composer-ic1)
← retract(X) ∧ ¬db(X)                        (composer-ic2)
```

Figure 4: A meta-theory for Example 9.

how some of the most preferred solutions for the meta-theory above are computed.

We follow one branch in the refutation tree, starting from the initial state $(\mathcal{G}_0, ST_0)$, where the initial set of goals is $\mathcal{G}_0 = \{$'true', **ic1**, **ic2**, **composer-ic1**, **composer-ic2**$\}$, and the initial store is $ST_0 = (\emptyset, \emptyset, \emptyset)$. Suppose that the first selected formula is

$F_1 \;=\; \textbf{ic1} \;= \leftarrow \texttt{fact}(\texttt{teaches}(\texttt{X},\texttt{Y})) \wedge \texttt{fact}(\texttt{teaches}(\texttt{X},\texttt{Z})) \wedge (\texttt{Y} \neq \texttt{Z}).$

Then, by D.2,

$$\begin{aligned}\mathcal{G}_1 \;=\; & \mathcal{G}_0 \setminus F_1 \;\cup \\ & \{ \leftarrow \texttt{db}(\texttt{teaches}(\texttt{X},\texttt{Y})) \wedge \neg\texttt{retract}(\texttt{teaches}(\texttt{X},\texttt{Y})) \wedge \texttt{fact}(\texttt{teaches}(\texttt{X},\texttt{Z})) \wedge (\texttt{Y} \neq \texttt{Z}), \\ & \leftarrow \texttt{insert}(\texttt{teaches}(\texttt{X},\texttt{Y})) \wedge \texttt{fact}(\texttt{teaches}(\texttt{X},\texttt{Z})) \wedge (\texttt{Y} \neq \texttt{Z}) \},\end{aligned}$$

and $ST_1 = ST_0$. Now, pick

$F_2 \;= \leftarrow \texttt{db}(\texttt{teaches}(\texttt{X},\texttt{Y})) \wedge \neg\texttt{retract}(\texttt{teaches}(\texttt{X},\texttt{Y})) \wedge \texttt{fact}(\texttt{teaches}(\texttt{X},\texttt{Z})) \wedge (\texttt{Y} \neq \texttt{Z}).$

Select `db(teaches(X,Y))`, unfold all the corresponding atoms in the database, and then, again by D.2, followed by E.2 and E.3,

$$\begin{aligned}\mathcal{G}_2 \;=\; & \mathcal{G}_1 \setminus F_2 \;\cup \\ & \{ \leftarrow \neg\texttt{retract}(\texttt{teaches}(\texttt{c}_1,\texttt{n}_1)) \wedge \texttt{fact}(\texttt{teaches}(\texttt{c}_1,\texttt{Z})) \wedge (\texttt{n}_1 \neq \texttt{Z}), \\ & \leftarrow \neg\texttt{retract}(\texttt{teaches}(\texttt{c}_2,\texttt{n}_2)) \wedge \texttt{fact}(\texttt{teaches}(\texttt{c}_2,\texttt{Z})) \wedge (\texttt{n}_2 \neq \texttt{Z}), \\ & \leftarrow \neg\texttt{retract}(\texttt{teaches}(\texttt{c}_2,\texttt{n}_3)) \wedge \texttt{fact}(\texttt{teaches}(\texttt{c}_2,\texttt{Z})) \wedge (\texttt{n}_3 \neq \texttt{Z}) \},\end{aligned}$$

and still $ST_2 = ST_1$. Pick then the second denial among the new goals that were added to $\mathcal{G}_2$. Denote this denial $F_3$. Since $F_3$ starts with a negated literal, N.2 applies, and the derivation process splits here to two branches. The second branch contains

$$\mathcal{G}_3 = \mathcal{G}_2 \setminus F_3 \cup \{ \leftarrow \texttt{retract}(\texttt{teaches}(\texttt{c}_2, \texttt{n}_2)), \leftarrow \texttt{fact}(\texttt{teaches}(\texttt{c}_2, \texttt{Z})) \wedge (\texttt{n}_2 \neq \texttt{Z}) \},$$

and still $ST_3 = ST_2$. Choose now the first new goal, i.e.,

$$F_4 = \leftarrow \texttt{retract}(\texttt{teaches}(\texttt{c}_2, \texttt{n}_2)).$$

Now, since $\Delta_3 = \emptyset$, the only option is to add $F_4$ to $\Delta_3^*$. Thus, by A.2,

$$\mathcal{G}_4 = \mathcal{G}_3 \setminus F_4 \text{ and } ST_4 = (\emptyset, \{F_4\}, \emptyset).$$

Assume, now, that we take the second new goal of $\mathcal{G}_3$:

$$F_5 = \leftarrow \texttt{fact}(\texttt{teaches}(\texttt{c}_2, \texttt{Z})) \wedge (\texttt{n}_2 \neq \texttt{Z}).$$

Following a similar process of unfolding data as described above, using $\texttt{db}(\texttt{teaches}(\texttt{c}_2, \texttt{n}_3))$, we end-up with

$$\leftarrow \texttt{retract}(\texttt{teaches}(\texttt{c}_2, \texttt{n}_3)) \wedge (\texttt{n}_2 \neq \texttt{n}_3).$$

Selecting the negative literal $(\texttt{n}_2 \neq \texttt{n}_3)$, N.2 applies again. The first branch quickly results in failure after adding $(\texttt{n}_2 = \texttt{n}_3)$ to $\mathcal{E}$. The second branch adds $\leftarrow (\texttt{n}_2 = \texttt{n}_3)$ and $\texttt{retract}(\texttt{teaches}(\texttt{c}_2, \texttt{n}_3))$ to the set of goals. The former one is added to the constraint store, as $(\texttt{n}_2 \neq \texttt{n}_3)$, and simplifies to true. Assume the latter is selected next. Let this be the $i$-th step. We have that by now $\Delta_{i-1}$ (the set of abducible predicates produced until the current step) is empty, thus the only option is to abduce $\texttt{retract}(\texttt{teaches}(\texttt{c}_2, \texttt{n}_3))$. Thus, by A.1, $ST_i$ consists of:

$$\Delta_i = \{\texttt{retract}(\texttt{teaches}(\texttt{c}_2, \texttt{n}_3))\}, \quad \Delta_i^* = \{F_4\} = \{\leftarrow \texttt{retract}(\texttt{teaches}(\texttt{c}_2, \texttt{n}_2))\},$$
$$\mathcal{E}_i = \mathcal{E}_{i-1}, \text{ and}$$
$$\mathcal{G}_i = \mathcal{G}_{i-1} \setminus \{\texttt{retract}(\texttt{teaches}(\texttt{c}_2, \texttt{n}_3))\} \cup \{\leftarrow \texttt{teaches}(\texttt{c}_2, \texttt{n}_2) = \texttt{teaches}(\texttt{c}_2, \texttt{n}_3)\}.$$

As the last goal is certainly satisfied, **ic1** is resolved in this branch.

Now we turn to **ic2**. So:

$$F_{i+1} = \textbf{ic2} = \leftarrow \texttt{fact}(\texttt{teacher}(\texttt{X})) \wedge \neg \texttt{gives_courses}(\texttt{X}).$$

The evaluation of $F_{i+1}$ for either $x = n_1$ or $x = n_2$ is successful, so the only interesting case is when $x = n_3$. In this case the evaluation leads to the goal `gives_courses(X)`. Unfolding this goal yields that $\texttt{fact}(\texttt{teaches}(\texttt{Y}, \texttt{n}_3))$ appears in the goal set. In order to satisfy this goal, it should be resolved either with one of the composer's rules (using D.1). The first rule (i.e., `fact(X) ← db(X) ∧¬ retract(X))` leads to a failure (since $\texttt{retract}(\texttt{teaches}(\texttt{c}_2, \texttt{n}_3))$ is already in $\Delta$), and so the second rule of the composer, `fact(X) ← insert(X)`, must be applied. This leads to the abduction of $\texttt{insert}(\texttt{teaches}(\texttt{Y}, \texttt{n}_3))$. By **ic1**, $\texttt{Y} \neq \texttt{c}_1$ and $\texttt{Y} \neq \texttt{c}_2$ is derived[24]. Also, **composer-ic1** and **composer-ic2** are satisfied by the current state, so eventually the solution state that is reached from the derivation path described here,

24. One can verify that these constraints are indeed detected during the derivations process. We omit the details here in order to keep this example tractable.

contains the following sets:

$$\Delta = \{\texttt{retract}(\texttt{teaches}(\texttt{c}_2, \texttt{n}_3)),\ \texttt{insert}(\texttt{teaches}(\texttt{Y}, \texttt{n}_3))\},$$
$$\mathcal{E} = \{\texttt{Y} \neq \texttt{c}_1, \texttt{Y} \neq \texttt{c}_2\},$$

which means retraction of $\texttt{teaches}(\texttt{c}_2, \texttt{n}_3)$ and insertion of $\texttt{teaches}(\texttt{Y}, \texttt{n}_3)$ for some $\texttt{Y} \neq \texttt{c}_1$ and $\texttt{Y} \neq \texttt{c}_2$. The other solutions are obtained in a similar way.

**Note 2** Below are some remarks on the above derivation process.

1. The solution above contains a non-ground abducible predicate. This indeed is the expected result, since this solution resolves the contradiction with the integrity constraint **ic1** by removing the assumption that teacher $n_3$ teaches course $c_2$. As a result, teacher $n_3$ does not teach any course. Thus, in order to assure the other integrity constraint (**ic2**), the solution indicates that $n_3$ must teach *some* course (other than $c_1$ and $c_2$).

2. One possible (and realistic) explanation for the cause of the inconsistency in the database of Example 9 and Figure 4, is a typographic error. It might happen, for instance, that $\texttt{c}_2$ was mistakenly typed instead of, say, $\texttt{c}_3$, in $\texttt{teaches(c}_2\texttt{,n}_3\texttt{)}$. In this case, the database repair computed above pinpoints this possibility (in our case, then, $\texttt{Y}$ should be equal to $\texttt{c}_3$)[25]. This explanation cannot be explicitly captured, unless particular repairs with non-ground solutions are constructed, as indeed is the case here. While some other approaches that have been recently introduced (e.g., Bravo & Bertossi, 2003; Cali, Lembo, & Rosati, 2003) properly capture cases such as those of Example 9, to the best of our knowledge, no other *application* of database integration has this ability.

3. Once the system finds a solution that corresponds to a goal state $\mathcal{S}_g = (\mathcal{G}_g, ST_g)$ with $\mathcal{G}_g = \emptyset$ and $ST_g = (\Delta_g, \Delta^*_g, \mathcal{E}_g)$, the $\leq_i$-optimizer may be used such that whenever a state $\mathcal{S} = (\mathcal{G}_s, (\Delta_s, \Delta^*_s, \mathcal{E}_s))$ is reached, and $|\Delta_g| < |\Delta_s|$, the corresponding branch of the tree is pruned[26].

### 4.5 Soundness and Completeness

In this section we give some soundness and completeness results for the $\mathcal{A}$system, and relate these results to the model-based preferential semantics, considered in Section 3.

In what follows we denote by $\mathcal{T}$ an abductive meta-theory (constructed as described in Section 4.2) for composing $n$ given databases $\mathcal{DB}_1, \ldots, \mathcal{DB}_n$. Let also $\texttt{Proc}_{\texttt{ALP}}$ be some sound abductive proof procedure for $\mathcal{T}$[27]. The following proposition shows that $\texttt{Proc}_{\texttt{ALP}}$ provides a coherent method for integrating the databases that are represented by $\mathcal{T}$.

**Proposition 11** *Every abductive solution that is obtained by* $\texttt{Proc}_{\texttt{ALP}}$ *for the query '*`true`*' on a theory* $\mathcal{T}$*, is a repair of* $\mathcal{UDB}$.

25. The variable $\texttt{Y}$ is free and $\{\texttt{Y}/\texttt{c}_3\}$ is an answer substitution as it grounds $\Delta$ and satisfies $\mathcal{E}$.
26. As the size of $\Delta_s$ can only increase along the derivation, the state $\mathcal{S}$ cannot lead to a solution that is better than the one induced by $\mathcal{S}_g$, and so the corresponding branch of the tree can indeed be pruned.
27. That is, $\texttt{Proc}_{\texttt{ALP}}$ is a process for computing only the abductive solutions of $\mathcal{T}$, in the sense of Definition 14.

*Proof:* By the construction of $\mathcal{T}$ it is easy to see that all the conditions that are listed in Definition 5 are satisfied. Indeed, the first two conditions are assured by the integrity constraints of the composer. The last condition is also met since by the soundness of $\texttt{Proc}_{\texttt{ALP}}$ it produces abductive solutions $\Delta_i$ for a query '`true`' on $\mathcal{T}$. Thus, by the second property in Definition 14, for every such solution $\Delta_i = (\mathsf{Insert}_i, \mathsf{Retract}_i)$ we have that $\mathcal{P} \cup \Delta_i \models \mathcal{IC}$. Since $\mathcal{P}$ contains a data section with all the facts, it follows that $\mathcal{D} \cup \Delta_i \models \mathcal{IC}$, i.e. every integrity constraints follows from $\mathcal{D} \cup \mathsf{Insert}_i \setminus \mathsf{Retract}_i$. □

As SLDNFA is a sound abductive proof procedure (Denecker & De Schreye, 1998), it can be taken as the procedure $\texttt{Proc}_{\texttt{ALP}}$, and so Proposition 11 provides a soundness theorem for the current implementation of the $\mathcal{A}$system. When an optimizer is incorporated in the $\mathcal{A}$system, we have the following soundness result for the extended system:

**Theorem 1** (Soundness) *Every output that is obtained by the query '`true`' on $\mathcal{T}$, where the $\mathcal{A}$system is executed with a $\leq_c$-optimizer [respectively, with an $\leq_i$-optimizer], is a $\leq_c$-preferred repair [respectively, an $\leq_i$-preferred repair] of $\mathcal{UDB}$.*

*Proof:* Follows from Proposition 11 (since the $\mathcal{A}$system is based on SLDNFA which is a sound abductive proof procedure), and the fact that the $\leq_c$-optimizer prunes paths that lead to solutions that are not $\leq_c$-preferable. Similar arguments hold for systems with an $\leq_i$-optimizer. □

**Proposition 12** *Suppose that the query '`true`' has a finite SLDNFA-tree w.r.t. $\mathcal{T}$. Then every $\leq_c$-preferred repair and every $\leq_i$-preferred repair of $\mathcal{UDB}$ is obtained by running $\mathcal{T}$ in the $\mathcal{A}$system.*

*Outline of proof:* The proof that all the abductive solutions with minimal cardinality are obtained by the system is based on Theorem 10.1 of Denecker & De Schreye, 1998, where it is shown that SLDNFA$^o$, which is an extension of SLDNFA, aimed for computing solutions with minimal cardinality, is complete; see Denecker & De Schreye, 1998, Section 10.1, for further details. Similarly, the proof that all the abductive solutions which are minimal w.r.t. set inclusion are obtained by the system is based on Theorem 10.2 of Denecker & De Schreye, 1998, that shows that SLDNFA$_+$, which is another extension of SLDNFA, aimed for computing minimal solutions w.r.t. set inclusion, is also complete; see Denecker & De Schreye, 1998, Section 10.2, for further details.

Now, the $\mathcal{A}$system is based on the combination of SLDNFA$^o$ and SLDNFA$_+$. Moreover, as this system does not change the refutation tree (but only controls the way rules are selected), Theorems 10.1 and 10.2 in Denecker and De Schreye (1998) are applicable in our case as well. Thus, all the $\leq_c$- and the $\leq_i$-minimal solutions are produced. This in particular means that every $\leq_c$-preferred repair as well as every $\leq_i$-preferred repair of $\mathcal{UDB}$ is produced by our system. □

It should be noted that the last proposition does not guarantee that non-preferred repairs will not be produced (as this is not true in general). However, as the following theorem shows, the use of an optimizer excludes this possibility.

**Theorem 2** (Completeness) *In the notations of Proposition 12 and under its assumptions, the output of the execution of $\mathcal{T}$ in the $\mathcal{A}$system together with a $\leq_c$-optimizer [respectively, together with an $\leq_i$-optimizer] is exactly $!(\mathcal{UDB}, \leq_c)$ [respectively, $!(\mathcal{UDB}, \leq_i)$].*

*Proof:* We shall show the claim for the case of $\leq_c$; the proof w.r.t. $\leq_i$ is similar.

Let $(\mathsf{Insert}, \mathsf{Retract}) \in !(\mathcal{UDB}, \leq_c)$. By Proposition 12, $\Delta = (\mathsf{Insert}, \mathsf{Retract})$ is one of the solutions produced by the $\mathcal{A}$system for $\mathcal{T}$. Now, during the execution of the system together with the $\leq_c$-optimizer, the path that corresponds to $\Delta$ cannot be pruned from the refutation tree, since by our assumption $(\mathsf{Insert}, \mathsf{Retract})$ has a minimal cardinality among the possible solutions, so the pruning condition is not satisfied. Thus $\Delta$ will be produced by the $\leq_c$-optimized system. For the converse, suppose that $(\mathsf{Insert}, \mathsf{Retract})$ is some repair of $\mathcal{UDB}$ that is produced by the $\leq_c$-optimized system. Suppose for a contradiction that $(\mathsf{Insert}, \mathsf{Retract}) \notin !(\mathcal{UDB}, \leq_c)$. By the proof of Proposition 12, there is some $\Delta' = (\mathsf{Insert}', \mathsf{Retract}') \in !(\mathcal{UDB}, \leq_c)$ that is constructed by the $\mathcal{A}$system for $\mathcal{T}$, and $(\mathsf{Insert}', \mathsf{Retract}') <_c (\mathsf{Insert}, \mathsf{Retract})$. But $|\Delta'| < |\Delta|$, and so the $\leq_c$-optimizer would prune the path of the $\Delta$ solution once its cardinality becomes bigger than $|\Delta'|$. This contradicts our assumption that $(\mathsf{Insert}, \mathsf{Retract})$ is produced by the $\leq_c$-optimized system. □

**Note 3** The SLDNFA-resolution on which the $\mathcal{A}$system is based is an extension of SLDNF-resolution (Lloyd, 1987) and coincides with it for logic programs with empty sets of abducible predicates. SLDNF-resolution is complete only if its computation always terminates. SLDNFA inherits this property. This is the reason why the condition of a finite SLDNFA-tree is imposed in Proposition 12 and Theorem 2. Like SLDNF, the termination of SLDNFA can be guaranteed by imposing syntactic conditions on the program. We refer to (Verbaeten, 1999), where some conditions are proposed to guarantee the existence of a finite SLDNFA-tree.

In the context of our paper, floundering would arise in the presence of unsafe integrity constraints (e.g., $\forall x\, p(x)$). One way to eliminate this problem is to use a unary domain predicate *dom*, ranging over the objects of the database, and to add a range for each quantified variable in the integrity constraints, so that we obtain formulae of the form $\forall x(dom(x) \rightarrow \psi(x))$ and $\exists x(dom(x) \wedge \psi(x))$.

The following results immediately follow from the propositions above and those of Section 3 (unless explicitly said, the $\mathcal{A}$system is without optimizer).

**Corollary 1** *Suppose that the query '`true`' has a finite SLDNFA refutation tree w.r.t. input theory $\mathcal{T}$. Then:*

1. *for every output* $(\mathsf{Insert},\ \mathsf{Retract})$ *of the $\mathcal{A}$system there is a classical model $M$ of $\mathcal{IC}$ s.t.* $\mathsf{Insert} = M^t \setminus \mathcal{D}$ *and* $\mathsf{Retract} = \mathcal{D} \setminus M^t$.

2. *for every output* $(\mathsf{Insert},\ \mathsf{Retract})$ *of the $\mathcal{A}$system there is a 3-valued model $N$ of $\mathcal{D} \cup \mathcal{IC}$ s.t.* $\mathsf{Insert}^N = \mathsf{Insert}$ *and* $\mathsf{Retract}^N = \mathsf{Retract}$.

**Corollary 2** *In the notations of Corollary 1 and under its assumption, we have that:*

1. *for every output* (Insert, Retract) *that is obtained by running the $\mathcal{A}$system together with an $\leq_i$-optimizer [respectively, together with a $\leq_c$-optimizer], there is an $\leq_i$-maximally consistent element [respectively, a $\leq_c$-maximally consistent element] $N$ in $\mathcal{M}^{\mathcal{UDB}}$ s.t.* $\mathsf{Insert}^N = \mathsf{Insert}$ *and* $\mathsf{Retract}^N = \mathsf{Retract}$.

2. *for every $\leq_i$-maximally consistent element [respectively, $\leq_c$-maximally consistent element] $N$ in $\mathcal{M}^{\mathcal{UDB}}$ there is a solution* (Insert, Retract) *that is obtained by running the $\mathcal{A}$system together with an $\leq_i$-optimizer [respectively, together with a $\leq_c$-optimizer] s.t.* $\mathsf{Insert} = \mathsf{Insert}^N$ *and* $\mathsf{Retract} = \mathsf{Retract}^N$.

The last corollaries show that the operational semantics, induced by the $\mathcal{A}$system, can also be represented by a preferential semantics, in terms of preferred models of the theory. The set $\mathcal{R}(\mathcal{UDB}, \leq)$ that represents the intended meaning of how to '$\leq$-recover' the database $\mathcal{UDB}$, can therefore be obtained computationally, by the set

$$\{(\mathsf{Insert}, \mathsf{Retract}) \mid (\mathsf{Insert}, \mathsf{Retract}) \text{ is an output of the } \mathcal{A}\text{system with an } \leq\text{-optimizer}\},$$

or, equivalently, it can be described in terms of preferred models of the theory, by the following set:

$$\{(\mathsf{Insert}^N, \mathsf{Retract}^N) \mid N \text{ is } \leq\text{-maximally consistent in } \mathcal{M}^{\mathcal{UDB}}\}.$$

### 4.6 Handling Specialized Information

The purpose of this section is to demonstrate the potential usage of our system in more complex scenarios, where various kinds of specialized data are incorporated in the system. In particular, we briefly consider time information and source identification. We also give some guidelines on how to extend the system with capabilities of handling these kinds of information.

#### 4.6.1 Timestamped Information

Many database applications contain temporal information. This kind of data may be divided in two types: time information that is part of the data itself, and time information that is related to database operations (e.g., records on database update time). Consider, for instance, $\text{birth_day(John,15/05/2001)}_{\mathsf{16/05/2001}}$. Here, John's date of birth is an instance of the former type of time information, and the subscripted data that describes the time in which this fact was added to the database, is an instance of the latter type of time information.

In our approach, timestamp information can be integrated by adding a temporal theory describing the state of the database at any particular time point. One way of doing so is by using *situation calculus*. In this approach a database is described by some initial information and a history of events performed during the database lifetime (see Reiter, 1995). Here we use a different approach, which is based on *event calculus* (Kowalski & Sergot, 1986). The idea is to make a distinction between two kinds of events, `add_db` and `del_db`, that describe the database modifications, and the composer-driven events `insert` and `retract` that are used for constructing database repairs. In this view, the extended composer has the following form:

```
holds_at(P,T) ← initially(P) ∧ ¬clipped(0,P,T)
holds_at(P,T) ← add(P,E) ∧ E<T ∧ ¬clipped(E,P,T)

clipped(E,P,T) ← del(P,C) ∧ E≤C, C<T

add(P,T) ← add_db(P,T)
add(P,T) ← insert(P,T)
del(P,T) ← del_db(P,T)
del(P,T) ← retract(P,T)

← insert(P,T) ∧ retract(P,T)
← insert(P,T) ∧ add_db(P,T)
← retract(P,T) ∧ del_db(P,T)
```

Note that in the above extended representation, the integrity constraints must be carefully specified. Consider, e.g. the statement that a person can be born only on one date:

```
← holds_at(birth_day(P,D1),T) ∧ holds_at(birth_day(P,D2),T) ∧ D1≠D2
```

The problem here is that to ensure consistency, this constraint must be checked at every point in time. This may be avoided by a simple rewriting that ensures that the constraint will be verified only when an event for that person occurs:

```
ic(P,T) ← holds_at(birth_day(P,D1),T) ∧ holds_at(birth_day(P,D2),T) ∧ D1≠D2
← add_db(birth_day(P,_),T) ∧ NT = T+1 ∧ ic(P,NT)
← insert(birth_day(P,_),T) ∧ NT = T+1 ∧ ic(P,NT)
← ic(P,0)
```

**Note 4** In the last example we have used *temporal integrity constraints* in order to resolve contradicting update events. Clearly, contradicting events do not necessarily yield a classically inconsistent database, and so the role of such integrity constraints is to express possible events in terms of time and causation, and – if necessary – describe their consequence as a violation of consistency.

Instead of using temporal integrity constraints and event calculus, one could repair a database with time-stamps by using some time-based criterion for making preferences among its repairs. For instance, denote by $\mathtt{db}(x_1, \ldots, x_n)_{\mathtt{t}}$ that the data-fact $\mathtt{db}(x_1, \ldots, x_n)$ has a timestamp $\mathtt{t}$, and suppose that $(\mathsf{Insert}, \mathsf{Retract})$ and $(\mathsf{Insert}', \mathsf{Retract}')$ are two repairs of a database $(\mathcal{D}, \mathcal{IC})$. A time-based criterion for preferring $(\mathsf{Insert}, \mathsf{Retract})$ over $(\mathsf{Insert}', \mathsf{Retract}')$ could state, e.g., that for every data-fact $\mathtt{db}(x_1, \ldots, x_n)$ and timestamps $\mathtt{t}_1, \mathtt{t}_2$ s.t. $\mathtt{db}(x_1, \ldots, x_n)_{\mathtt{t}_1}$ follows from $\mathcal{D} \cup \mathsf{Insert} \setminus \mathsf{Retract}$ and $\mathtt{db}(x_1, \ldots, x_n)_{\mathtt{t}_2}$ follows from $\mathcal{D} \cup \mathsf{Insert}' \setminus \mathsf{Retract}'$, necessarily $\mathtt{t}_1 \geq \mathtt{t}_2$. A more detailed treatment of this issue is outside the scope of this paper.

The interested reader may refer, e.g., to (Sripada, 1995; Mareco & Bertossi, 1999) for a detailed discussion on the use of logic programming based approaches to the specification of temporal databases. Such specifications can be easily combined with those for repairs, given above.

### 4.6.2 Keeping Track of Source Identities

There are cases in which it is important to preserve the identity of the database from which a specific piece of information was originated. This is useful, for instance, when one wants to make preferences among different sources, or when some specific source should be filtered out (e.g, when the corresponding database is not available or becomes unreliable). This kind of information may be decoded by adding another argument to every fact, which denotes the identity of its origin. This requires minor modifications in the basic composer, since the composer controls the way in which the data is integrated. As such, it is the only component that can keep track on the source of the information.

Suppose, then, that for every database fact we add another argument that identifies its source. I.e., `db(X,S)` denotes that `X` is a fact originated from a database `S`. The composer then has the following form:

```
fact(X,S) ← db(X,S) ∧ ¬retract(X)
fact(X,composer) ← insert(X)
← insert(X) ∧ db(X,S)
← retract(X) ∧ ¬db(X,S)
```

Note that the composer considers itself as an extra source that inserts brand new data facts. Now it is possible, e.g., to trace information that comes from a specific source, make preferences among different sources (by specifying appropriate integrity constraints), and filter data that comes from certain sources. The last property is demonstrated by the next rule:

```
validFact(X) ← fact(X,S) ∧ trusted_source(S)
```

where `trusted_source` enumerates all reliable sources of the data.

Note that the last example of 'source identification' can be further extended in order to make preferences among different sources (and not only ignoring some unreliable sources). By introducing a new predicate, `trust(Source,Amount)`, that attaches a certain level of reliability to each source, it is possible, in case of conflicts, to prefer sources with higher reliability as follows:

$$\leftarrow \texttt{fact(X,S)} \wedge \texttt{db(X,S}_0\texttt{)} \wedge \texttt{S} \neq \texttt{S}_0 \wedge \texttt{more_trusted(S}_0\texttt{,S)}$$
$$\texttt{more_trusted(S}_0\texttt{,S)} \leftarrow \texttt{trust(S}_0\texttt{,A}_0\texttt{)} \wedge \texttt{trust(S,A)} \wedge \texttt{A}_0 > \texttt{A}$$

This method is particularly useful when the integrity constraint above acts as a functional dependency on specific facts. The following example (originally introduced in Subrahmanian, 1994) demonstrates this.

**Example 10** Consider the following simple scenario of 'target recognition', where three sensors of an autonomous vehicle, which have different degrees of reliability, should identify objects in the vehicle's neighborhood:

```
trust(radar,10)
trust(gunchar,8)
```

```
trust(speedometer,5)

db(observe(object1,t72),radar)
db(observe(object1,t60),gunchar)
db(observe(object1,t80),speedometer)
```

$\leftarrow$ `fact(observe(O,V`$_1$`),S)` $\wedge$ `db(observe(O,V`$_2$`),S`$_0$`)` $\wedge$ `S`$\neq$`S`$_0$ $\wedge$ `more_trusted(S`$_0$`,S)`

As the radar has the highest reliability, its observation will be preserved. The observations of the other sensors will be retracted from the database.

## 5. Discussion and an Overview of Related Works

The interest in systems for coherent integration of databases has been continuously growing in the last few years (see, e.g, Olivé, 1991; Baral et al., 1991, 1992; Revesz, 1993; Subrahmanian, 1994; Bry, 1997; Gertz & Lipeck, 1997; Messing, 1997; Lin & Mendelzon, 1998; Liberatore & Schaerf, 2000; Ullman, 2000; Greco & Zumpano, 2000; Greco et al., 2001; Franconi et al., 2001; Lenzerini, 2001, 2002; Arenas et al., 1999, 2003; Bravo & Bertossi, 2003; Cali et al., 2003, and many others). Already in the early works on this subject it became clear that the design of systems for data integration is a complex task, which demands solutions to many questions from different disciplines, such as belief revision, merging and updating, reasoning with inconsistent information, constraint enforcement, query processing and – of course – many aspects of knowledge representation. In this section we shall address some of these issues.

One important aspect of data integration systems is how concepts in the independent (stand-alone) data-sources and those of the unified database are mapped to each other. A proper specification of the relations between the source schemas and the schema of the amalgamated data exempts the potential user from being aware where and how data is arranged in the sources. One approach for this mapping, sometimes called *global-centric* or *global-as-view* (Ullman, 2000), requires that the unified schema should be expressed in terms of the local schemas. In this approach, every term in the unified schema is associated with a view (alternatively, a query) over the sources. This approach is taken by most of the systems for data integration, as well as ours. The main advantage of this approach is that it induces a simple query processing strategy that is based on unfolding of the query, and uses the same terminology as that of the databases. This indeed is the case in the abductive derivation process, defined in Section 4.3.1. The other approach, sometimes called *source-centric* or *local-as-view* (used, e.g., in Bertossi et al., 2002), considers every source as a view over the integrated database, and so the meaning of every source is obtained by concepts of the global database. In particular, the global schema is independent of the distributed ones. This implies, in particular, that an addition of a new source to the system requires only to provide local definitions and not necessarily involves changes in the global schema. The main advantage of the latter approach is, therefore, that it provides a better setting for maintenance. For a detailed discussion on this topic, see (Ullman, 2000; Lenzerini, 2001; Cali et al., 2002; Van Nuffelen et al., 2004). More references and a survey on different approaches to data integration appear in the papers of Batini, Lenzerini, and Navathe (1986),

Rahm and Bernstein (2001), and Lenzerini (2002).

Another major issue that has to be addressed is the ability of data integration systems to properly cope with dynamically evolving worlds. In particular, the domain of discourse should not be fixed in advance, and information may be revised on a regular basis. The last issue is usually handled by methods of *belief revision* (Alchourrón et al., 1995; Gärdenfors & Rott, 1995) and *nonmonotonic reasoning*. In the context of belief revision it is common to make a distinction between revisions of integrity constraints and changes in the sets of the data-facts, since the two types of information have different nature and thus may require different approaches for handling dynamic changes. When the set of integrity constraints is given in a clause form, methods of *dynamic logic programing* (Alferes et al., 2000, 2002) may be useful for handling revisions. As noted in (Alferes et al., 2002), assuming that each local database is consistent (as in our case), dynamic logic programing (together with a proper language for implementing it, like LUPS (Alferes et al., 2002)) provides a way of avoiding contradictory information, and so this may be viewed as a method of updating a database by a sequence of integrity constraints that arrive at different time points.

When the types of changes are predictable, or can be characterized in some sense, temporal integrity constraints (in the context of *temporal databases*) can be used in order to specify how to treat new information. This method is also useful when the revision criteria are known in advance (e.g., ‘in case of collisions, prefer the more recent data’, cf. Section 4.6.1). See, e.g., (Sripada, 1995; Mareco & Bertossi, 1999) for a detailed discussion on temporal integrity constraints and temporal databases in a logic programming based formalisms.

The second type of revisions (i.e., modifications of data-facts) is obtained here through the (preferred) repairs of the unified database, which induce corresponding modifications of data-facts. A repair is usually induced by a method of restoring (or assuring) consistency of the amalgamated database by a minimal amount of change. As in our case, the minimization criterion is often determined by the aspiration to remain ‘as close as possible’ to the set of the collective information. This is a typical kind of a *repair goal*, and the standard ways of formally expressing it are by enumeration methods, such as the following[28]:

- Minimizing the Hamming distance between the (propositional) models of the unified database and its repairs (Liberatore & Schaerf, 2000), or minimizing the distance between the corresponding three-valued interpretations (de Amo et al., 2002) according to a suitable generalization of Hamming distance.
- Minimizing the symmetric distance between the sets of consequences of the corresponding databases (Arenas, Bertossi, & Chomicki, 1999; Arenas, Bertossi, & Kifer, 2000; Bertossi, Chomicki, Cortés, & Gutierrez, 2002) or, equivalently, minimizations in terms of set inclusion (Greco & Zumpano, 2000).
- When the underlying data is prioritized, the corresponding quantitative information is also considered in the computations of distances (see, for instance, the work of Liberatore & Schaerf, 2000).

28. See also (Gertz & Lipeck, 1997, Section 5) for a discussion on repair strategies.

Various ways of computing (preferred/minimal) repairs are described in the literature, among which are proof-theoretical (deductive) methods (Bertossi & Schwind, 2002; de Amo et al., 2002), abductive methods (Kakas & Mancarella, 1990a; Inoue & Sakama, 1995; Sakama & Inoue, 1999, 2000), and algorithmic approaches that are based on computations of maximal consistent subsets (Baral et al., 1991, 1992), or use techniques from model-based diagnosis (Gertz & Lipeck, 1997). A common approach is to view a database as a logic program, and to adopt standard techniques of giving semantics to logic programs in order to compute database repairs. For instance, stable-model semantics on disjunctive logic programs is used for computing repairs in (Greco & Zumpano, 2000; Greco et al., 2001; Franconi et al., 2001; Arenas et al., 2003), and resolution-based procedures for integrating several annotated databases are introduced by Subrahmanian (1990, 1994). As it follows from Section 4, the application introduced here is also based on an extended resolution strategy, applied on logic programs that may contain negation-as-failure operators and abducible predicates.

As repairing a database means in particular elimination of contradictions, reasoning with inconsistent information has been a major challenge for data integration systems. First, it is important to note in this respect that not every formalism for handling inconsistency is acceptable in the context of databases, even if the underlying criterion for handling inconsistency is the same as one of the repair goals mentioned above. The following example demonstrates such a case:

**Example 11** (Arenas, Bertossi, & Chomicki, 1999) Consider the following (inconsistent) database: $\mathcal{DB} = (\{p,\ q\}, \{\neg(p \wedge q)\})$. In the approach of Lin (1996), for instance, $p \vee q$ may be inferred as the repaired database, following a strategy of minimal change. However, in this approach none of $p$, $q$, and $\neg(p \wedge q)$ holds in the repaired database. In particular (since in (Lin, 1996) there is no distinction between data-facts and integrity constraints), the integrity constraint $\{\neg(p \wedge q)\}$ itself cannot be inferred, which violates the intended meaning of an integrity constraint in databases.

Many techniques for consistency enforcement and repairs of constraint violations have been suggested, among which are methods for resolving contradictions by quantitative considerations, such as 'majority vote' (Lin & Mendelzon, 1998; Konieczny & Pino Pérez, 2002) or qualitative ones (e.g., defining priorities on different sources of information or preferring certain data over another, as in Benferhat, Cayrol, Dubois, Lang, & Prade, 1993, and Arieli, 1999). Another common method of handling inconsistent (and incomplete) information is by turning to multi-valued semantics. Three-valued formalisms such as the one considered in Section 3 are used as a semantical basis of paraconsistent methods to construct database repairs (de Amo, Carnielli, & Marcos, 2002) and are useful in general for pinpointing inconsistencies (Priest, 1991). Other approaches use lattice-based semantics to decode within the language itself some meta-information, such as confidence factors, amount of belief for or against a specific assertion, etc. These approaches combine corresponding formalisms of knowledge representation, such as annotated logic programs (Subrahmanian, 1990, 1994; Arenas et al., 2000) or bilattice-based logics (Fitting, 1991; Arieli & Avron, 1996; Messing, 1997), together with non-classical refutation procedures (Fitting, 1989; Subrahmanian, 1990; Kifer & Lozinskii, 1992) that allow to detect inconsistent parts of a database and maintain them.

## 6. Summary and Future Work

In this paper we have developed a formal declarative foundation for rendering coherent data, provided by different databases, and presented an application that implements this approach. Like similar applications (e.g., Subrahmanian, 1994; Bertossi, Arenas, & Ferretti, 1998; Greco & Zumpano, 2000; Liberatore & Schaerf, 2000), our system mediates among the sources of information and also between the reasoner and the underlying data.

Composition of several data-sources is encoded by meta-theories in the form of abductive logic programs, and it is possible to extend these theories by providing meta-information on the data-facts, such as time-stamps and source identities. Moreover, since the reasoning process of the system is based on a pure generalization of classical refutation procedures, no syntactical embedding of first-order formulae into other languages, nor any extension of two-valued semantics, is necessary.

Due the inherent modularity of the system, each component is independent and can be modified to meet different needs. Thus, for instance, the underlying solver may be replaced with any other solver that is capable of dealing with the meta-theory, and any improvement of the optimizer will affect the whole system and its efficiency, regardless the nature of its input. Also, the way of keeping data coherent is encapsulated in the component that integrates the data (i.e., the composer). This implies, in particular, that no input from the reasoner nor any other external policy for making preferences among conflicting sources is compulsory in order to resolve contradictions.

As we have shown, the operational semantics for inconsistent databases, induced by the $\mathcal{A}$system, is strongly related to (multi-valued) preferential semantics. As preferential semantics provides the background for many non-monotonic and paraconsistent formalisms (e.g., Shoham, 1988; Priest, 1989, 1991; Kifer & Lozinskii, 1992; Arieli & Avron, 1996; Arieli, 1999, 2003), this implies that the $\mathcal{A}$system may be useful for reasoning with general uncertain theories (not necessarily in the form of databases).

It is important to note that our composing system inherits the functionality of the underlying solver. The outcome of this is flexibility, modularity, simple interaction with different sources of information, and the ability to reason with *any* set of first-order formulae of integrity constraints[29]. To the best of our knowledge no other application of data integration has this ability.

There are several directions for further exploration. First, as we have already noted, two more phases, which have not been considered here, might be needed for a complete process of data integration:

29. Provided, of-course, that the constraints do not lead to floundering.

a) translation of difference concepts to a unified ontology, and

b) integration of integrity constraints.

So far, formalisms for dealing with the first item (e.g., Lenzerini, 2001, 2002; Van Nuffelen et al., 2004) mainly focus on the mutual relations between the global schema and the source (local) schemas, in particular how concepts of each ontology map to each other. On the other hand, formalisms for handling the second item concentrate on nonmonotonic reasoning for dynamically evolving (and mutually inconsistent) worlds. A synthesis of the main ideas behind these approaches, and incorporating them in our system, is a major challenge for future work.

Another important issue that deserves attention is the repair of inconsistency in the context of deductive databases with integrity constraints and definitions of predicates, often called *view predicates*. We refer to (Denecker, 2000) for a sketch on how this may be done. This kind of data may be further combined with (possibly inconsistent) temporal information, (partial) transactions, and (contradictory) update information.

Finally, since different databases may have different information about the same predicates, it is reasonable to use some weakened version of the closed word assumption as part of the integration process (for instance, an assumption that something is false unless it is in the database, or unless some other database has some information about it).

## Acknowledgements

We would like to thank the anonymous reviewers for many helpful comments and suggestions. This research was supported by the Research Fund K.U.Leuven and by FWO–Vlaanderen.